%% file: main.tex
\documentclass[letterpaper]{article} 
\usepackage{aaai25}  
\usepackage{times}  
\usepackage{helvet}  
\usepackage{courier}  
\usepackage[hyphens]{url}  
\usepackage{graphicx} 
\usepackage{natbib}  
\usepackage{caption} 
\usepackage{algorithm}
\usepackage{algorithmic}

\usepackage{amsmath}
\usepackage{amsfonts}
\usepackage{amssymb}
\usepackage{dsfont}
\usepackage{booktabs}

\usepackage{newfloat}
\usepackage{listings}
\DeclareCaptionStyle{ruled}{labelfont=normalfont,labelsep=colon,strut=off} 
\floatstyle{ruled}
\newfloat{listing}{tb}{lst}{}
\floatname{listing}{Listing}
\title{Segment Any 3D Gaussians}
\author{
    Jiazhong Cen\textsuperscript{\rm 1}, Jiemin Fang\textsuperscript{\rm 2}, Chen Yang\textsuperscript{\rm 1}, Lingxi Xie\textsuperscript{\rm 2}, Xiaopeng Zhang\textsuperscript{\rm 2}, Wei Shen\textsuperscript{\rm 1}\thanks{Corresponding author}, Qi Tian\textsuperscript{\rm 2}\\
}
\affiliations{
    \textsuperscript{\rm 1}MoE Key Lab of Artificial Intelligence, AI Institute, Shanghai Jiao Tong University\\
    \textsuperscript{\rm 2}Huawei Technologies Co., Ltd.\\
    \{jiazhongcen, ycyangchen, wei.shen\}@sjtu.edu.cn, \\
    \{jaminfong, 198808xc, zxphistory\}@gmail.com, tian.qi1@huawei.com
}

\begin{document}

\maketitle

\begin{abstract}
This paper presents SAGA (\underline{S}egment \underline{A}ny 3D \underline{GA}ussians), a highly efficient 3D promptable segmentation method based on 3D Gaussian Splatting (3D-GS). Given 2D visual prompts as input, SAGA can segment the corresponding 3D target represented by 3D Gaussians within \textbf{4 ms}. This is achieved by attaching an scale-gated affinity feature to each 3D Gaussian to endow it a new property towards multi-granularity segmentation. Specifically, a scale-aware contrastive training strategy is proposed for the scale-gated affinity feature learning. It 1) distills the segmentation capability of the Segment Anything Model (SAM) from 2D masks into the affinity features and 2) employs a soft scale-gate mechanism to deal with multi-granularity ambiguity in 3D segmentation through adjusting the magnitude of each feature channel according to a specified 3D physical scale. Evaluations demonstrate that SAGA achieves real-time multi-granularity segmentation with quality comparable to state-of-the-art methods. As one of the first methods addressing promptable segmentation in 3D-GS, the simplicity and effectiveness of SAGA pave the way for future advancements in this field.
\end{abstract}


\begin{links}
    \link{Code}{https://github.com/Jumpat/SegAnyGAussians}
\end{links}

\begin{figure*}[htbp]
    \centering
    \includegraphics[width=0.8\linewidth]{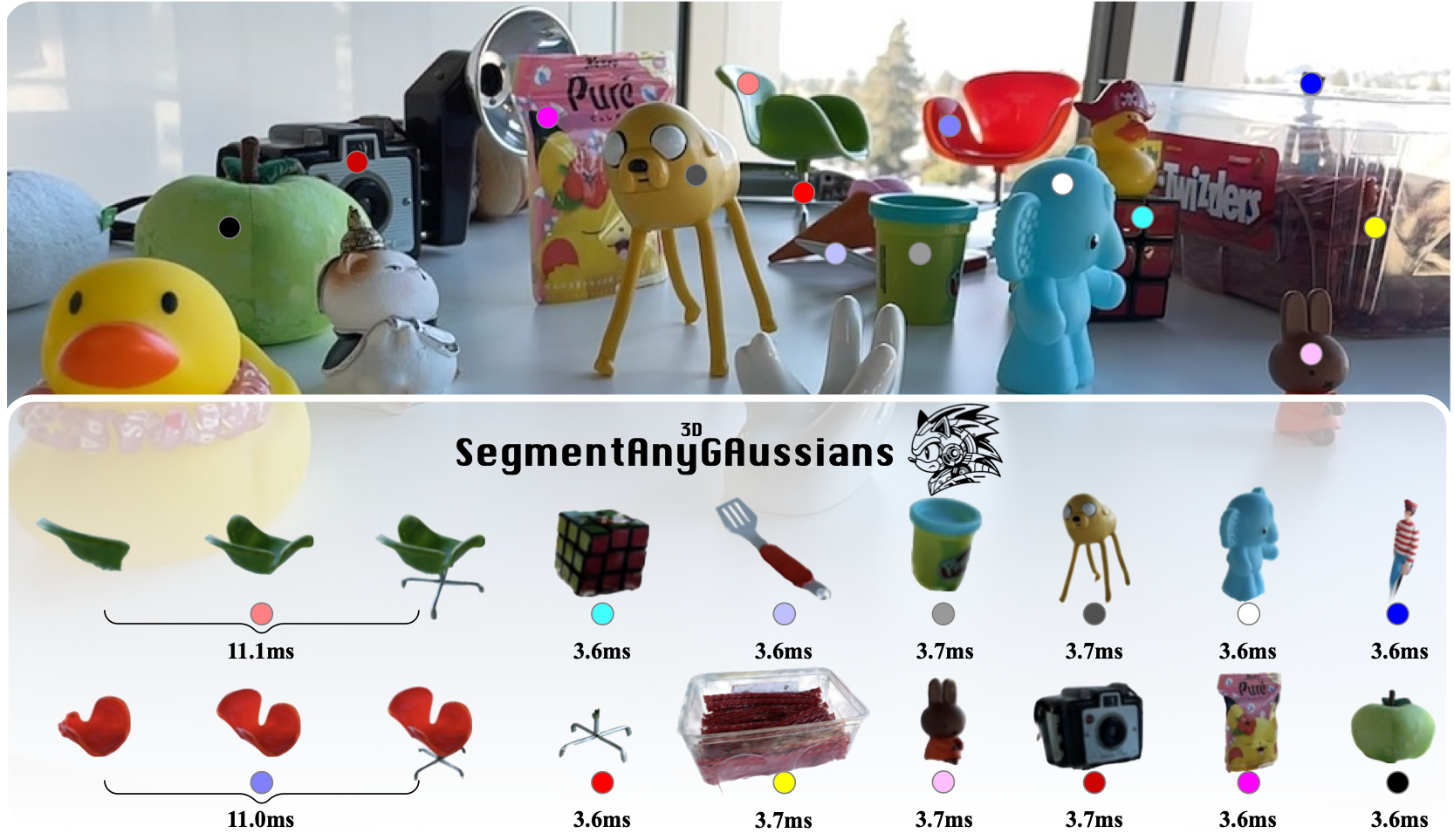}
    \caption{SAGA performs promptable multi-granularity segmentation within \textbf{milliseconds}. Prompts are marked by points.}
    \label{fig:teaser}
\end{figure*}

\section{Introduction}

Promptable segmentation has attracted increasing attention and has seen significant advancements, particularly with the development of 2D segmentation foundation models such as the Segment Anything Model (SAM)~\cite{sam}. However, 3D promptable segmentation remains relatively unexplored due to the scarcity of 3D data and the high cost of annotation. To address these challenges, many studies~\cite{sa3d, chen2023interactive, omniseg3d, garfield, nerfsos, samnerfhq} have proposed to extend SAM's 2D segmentation capabilities to 3D using radiance fields, achieving notable success.

In this paper, we focus on promptable segmentation in 3D Gaussian Splatting (3D-GS)~\cite{3dgs}, which represents a significant milestone in radiance fields research due to its superior rendering quality and efficiency compared to its predecessors. We highlight that, in contrast to previous radiance fields, the explicit 3D Gaussian structure is an ideal carrier for 3D segmentation, as segmentation capabilities can be integrated into 3D-GS as an intrinsic attribute, without necessitating an additional bulky segmentation module.

Accordingly, we propose SAGA (\underline{S}egment \underline{A}ny 3D \underline{GA}ussians), a 3D promptable segmentation method that integrates the segmentation capabilities of SAM into 3D-GS seamlessly. SAGA takes 2D visual prompts as input and outputs the corresponding 3D target represented by 3D Gaussians.
To achieve this purpose, two primary challenges are faced. First, SAGA should figure out an efficient way to endow each 3D Gaussian with the ability of 3D segmentation, so that the high efficiency of 3D-GS can be preserved. Second, as a robust promptable segmentation method, SAGA must effectively address multi-granularity ambiguity, where a single 3D Gaussian may belong to different parts or objects at varying levels of granularity.

To address the two challenges, SAGA respectively introduces two solutions. First, SAGA attaches an affinity feature to each 3D Gaussian in a scene to endow it with a new property towards segmentation. The similarity between two affinity features indicates whether the corresponding 3D Gaussians belong to the same 3D target. Second, inspired by GARField~\cite{garfield}, SAGA employs a soft scale-gate mechanism to handle multi-granularity ambiguity. Depending on a specified 3D physical scale, the scale gate adjusts the magnitude of each feature channel. This mechanism maps the Gaussian affinity features into different sub-spaces for various scales, thereby preserving the multi-granularity information and meanwhile mitigating the distraction in feature learning brought by multi-granularity ambiguity.
To realize the two solutions, SAGA proposes a scale-aware contrastive training strategy, which distills the segmentation capability of SAM from 2D masks into scale-gated affinity features. This strategy determines the correlations between a pair of pixels within an image based on 3D scales. These correlations are then used to supervise the rendered affinity features through a correspondence distillation loss. The correlation information is transmitted to the Gaussian affinity features via backpropagation facilitated by the differentiable rasterization algorithm. After training, SAGA achieves real-time multi-granularity segmentation precisely.

\section{Related Work}
\label{sec:related}

\paragraph{2D Promptable Segmentation} The task of 2D promptable segmentation is proposed by~\citet{sam}, which aims to return segmentation masks given input prompts that specify the segmentation target in an image. To address this problem, they introduce the Segment Anything Model (SAM), a groundbreaking segmentation foundation model. A similar model to SAM is SEEM~\cite{seem}, which also achieves competitive performance. Prior to these models, the most closely related task to promptable 2D segmentation is interactive image segmentation~\cite{interactive,grady2006random,gulshan2010geodesic, GrabCut,chen2022focalclick,sofiiuk2022reviving, liu2023simpleclick}. Inspired by the success of SAM, many studies~\cite{sam3d, sampro3d, guo2024samguidedgraphcut3d, yin2024sai3d} proposed to use SAM for 3D segmentation. Different from them, we focus on lifting the ability of SAM to 3D via 3D-GS.

\paragraph{3D Segmentation in Radiance Fields} 
With the success of radiance fields~\cite{NeRF, dvgo, tensorf, mipnerf360, InstantNGP, bakingnerf, plenoxel, nex, autoint, TiNeuVox}, numerous studies have explored 3D segmentation within them. \citet{semantic-nerf} proposed Semantic-NeRF, demonstrating the potential of Neural Radiance Field (NeRF) in semantic propagation and refinement. NVOS~\cite{nvos} introduced an interactive approach to select 3D objects from NeRF by training a lightweight MLP using custom-designed 3D features. By using 2D self-supervised models, approaches like N3F~\cite{n3f}, DFF~\cite{dff}, and ISRF~\cite{isrf} aim to elevate 2D visual features to 3D by training additional feature fields that can output 2D feature maps, imitating the original 2D features from different views. NeRF-SOS~\cite{nerfsos} and ContrastiveLift~\cite{contrastivelift} distill the 2D feature similarities into 3D features. There are also some other instance segmentation and semantic segmentation approaches~\cite{obsurf,giraffe,uorf,rfp, instance-nerf, dmnerf, panopticnerf, nesf, siddiqui2023panoptic} for radiance fields. Combined with CLIP~\cite{clip}, some approaches~\cite{lerf, 3dovs, n2f2, langsplat} proposed to conduct open-vocabulary 3D segmentation in radiance fields. With the popularity of SAM, a stream of studies~\cite{garfield, omniseg3d, gaussiangrouping, sa3d, gaga} proposed lifting the segmentation ability of SAM to 3D with radiance fields. SA3D~\cite{sa3d} adopts an iterative pipeline to refine the 3D mask grids with SAM. GaussianGrouping~\cite{gaussiangrouping} uses video tracking technology to align the inconsistent 2D masks extracted by SAM across different views and assigns labels to 3D Gaussians in a 3D-GS model with the aligned masks. OmniSeg3D~\cite{omniseg3d} employs a hierarchical contrastive learning method to automatically learn segmentation from multi-view 2D masks extracted by SAM. 

The approach most closely related to SAGA is GARField~\cite{garfield}, which addresses multi-granularity ambiguity in 3D segmentation using 3D physical scale, inspiring SAGA’s scale gate mechanism. However, GARField’s reliance on implicit feature fields for outputting 3D features requires repeated queries for segmentation at different scales, reducing efficiency. In contrast, the scale-gate mechanism of SAGA enhances efficiency by integrating directly with 3D-GS without additional computation.

\section{Method}
\label{sec:method}
In this section, we first give a brief review of 3D Gaussian Splatting (3D-GS)~\cite{3dgs} and the scale-conditioned 3D features~\cite{lerf, garfield}. Then we introduce the overall pipeline of SAGA, followed by explanation of the scale-gated Gaussian affinity features and the scale-aware contrastive learning.

\subsection{Preliminary}
\label{sec:preliminaries}
\paragraph{3D Gaussian Splatting (3D-GS)} Given a training dataset $\mathcal{I}$ of multi-view 2D images with camera poses, 3D-GS learns a set of 3D colored Gaussians $\mathcal{G} = \{\mathbf{g}_1, \mathbf{g}_2, ..., \mathbf{g}_{N}\}$, where $N$ denotes the number of 3D Gaussians in the scene. The mean of a Gaussian represents its position and the covariance indicates its scale. Accordingly, 3D-GS proposes a novel differentiable rasterization technology for efficient training and rendering. Given a specific camera pose, 3D-GS projects the 3D Gaussians to 2D and computes the color $\mathbf{C}(\mathbf{p})$ of a pixel $\mathbf{p}$ by blending a set of ordered Gaussians $\mathcal{G}_{\mathbf{p}}$ overlapping the pixel. Let $\mathbf{g}_i^{\mathbf{p}}$ denote the i-th Gaussian in $\mathcal{G}_{\mathbf{p}}$, this process is formulated as:
\begin{equation}
\label{eq:rasterization}
    \mathbf{C}(\mathbf{p})=\sum_{i = 1}^{|\mathcal{G}_{\mathbf{p}}|}\mathbf{c}_{\mathbf{g}_i^{\mathbf{p}}}\alpha_{\mathbf{g}_i^{\mathbf{p}}}\prod_{j=1}^{i-1}(1-\alpha_{\mathbf{g}_j^{\mathbf{p}}}),
\end{equation}
where $\mathbf{c}_{\mathbf{g}_i^{\mathbf{p}}}$ is the color of $\mathbf{g}_i^{\mathbf{p}}$ and $\alpha_{\mathbf{g}_i^{\mathbf{p}}}$ is given by evaluating the corresponding 2D Gaussian with covariance $\Sigma$ multiplied with a learned per-Gaussian opacity. 

\paragraph{Scale-Conditioned 3D Feature} 
LERF~\cite{lerf} first proposes the concept of a scale-conditioned feature field for learning from global image embeddings obtained from CLIP. GARField~\cite{garfield} then introduces it into the area of radiance field segmentation to tackle the multi-granularity ambiguity.
To compute the 3D mask scale $s_\mathbf{M}$ of a 2D mask $\mathbf{M}$, GARField projects $\mathbf{M}$ into 3D space with the camera intrinsic parameters and depth information predicted by a pre-trained radiance field. Let $\mathcal{P}$ denote the obtained point cloud, $\mathcal{X}(\mathcal{P}),\mathcal{Y}(\mathcal{P}),\mathcal{Z}(\mathcal{P})$ denote the set of 3D coordinate components of $\mathcal{P}$, the mask scale $s_\mathbf{M}$ is:
\begin{equation}
\label{eq:scale}
    s_\mathbf{M} = 2\sqrt{\texttt{std}(\mathcal{X}(\mathcal{P}))^2+\texttt{std}(\mathcal{Y}(\mathcal{P}))^2+\texttt{std}(\mathcal{Z}(\mathcal{P}))^2},
\end{equation}
where $\texttt{std}(\cdot)$ denotes the standard variation of a set of scalars. Since these scales are computed in 3D space, they are generally consistent across different views. SAGA uses the 3D scales for multi-granularity segmentation but realizes it in a more efficient way.

\subsection{Overall Pipeline}
\label{sec:overall}
\begin{figure*}
    \centering
    \includegraphics[width=0.9\linewidth]{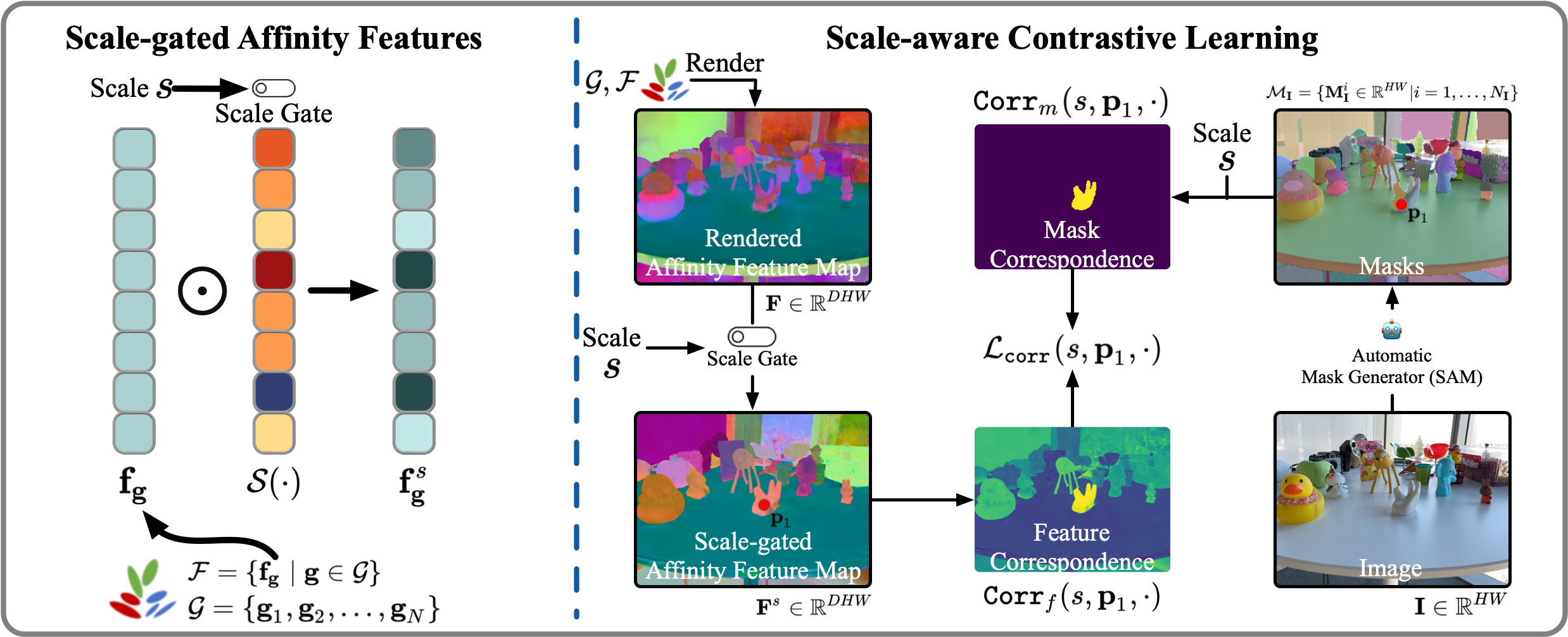}
    \caption{The architecture of SAGA. \textit{Left}: SAGA attaches a Gaussian affinity feature to each 3D Gaussian. The magnitude of different affinity feature channels are adjusted by a soft scale gate to handle multi-granularity ambiguity. \textit{Right}: SAGA distills segmentation ability of SAM into affinity features attached to 3D Gaussians in the 3D-GS model through scale-aware contrastive learning.}
    \label{fig:pipe}
\end{figure*}
The main components of SAGA are shown in Figure~\ref{fig:pipe}. Given a pre-trained 3D-GS model $\mathcal{G}$, 
SAGA attaches a Gaussian affinity feature $\mathbf{f}_\mathbf{g} \in \mathbb{R}^{D}$ for each 3D Gaussian $\mathbf{g}$ in $\mathcal{G}$. $D$ denotes the feature dimension. To handle the inherent multi-granularity ambiguity of 3D promptable segmentation, SAGA employs a soft scale gate mechanism to project these features into different scale-gated feature subspaces for various scales $s$.

To train the affinity features, SAGA extracts a set of multi-granularity masks $\mathcal{M}_{\mathbf{I}} = \{\mathbf{M}_\mathbf{I}^{i} \in \{0,1\}^{HW} | i = 1,...,N_\mathbf{I}\}$ for each image $\mathbf{I}$ in the training set $\mathcal{I}$ with SAM. $H,W$ denotes the height and width of $\mathbf{I}$ respectively. $N_\mathbf{I}$ is the number of extracted masks. For each mask $\mathbf{M}_\mathbf{I}^{i}$, its 3D physical scale $s_{\mathbf{M}_\mathbf{I}^{i}}$ is calculated using the depth predicted by $\mathcal{G}$ with the camera pose, as shown in Equation~\eqref{eq:scale}. Subsequently, SAGA employs a scale-aware contrastive learning strategy (Section~\ref{sec:training}) to distill the multi-granularity segmentation ability embedded in multi-view 2D masks into the scale-gated affinity features.
After training, at given scales, the affinity feature similarities between two Gaussians indicate whether they belong to the same 3D target. 

During inference (Section~\ref{sec:infer}), given a specific viewpoint, SAGA converts 2D visual prompts (points with scales) into corresponding 3D scale-gated query features to segment the 3D target by evaluating feature similarities with 3D affinity features. Additionally, with well-trained affinity features, 3D scene decomposition is achievable through simple clustering. Moreover, by integrating with CLIP, SAGA can perform open-vocabulary segmentation (see Section A.3 of the supplement) without requiring language fields.

\subsection{Gaussian Affinity Feature}
\label{sec:feature}
At the core of SAGA is the Gaussian affinity features $\mathcal{F} = \{\mathbf{f}_{\mathbf{g}} \mid \mathbf{g} \in \mathcal{G}\}$, which are learned from the multi-view 2D masks extracted by SAM.
To tackle the inherent multi-granularity ambiguity in promptable segmentation, inspired by GARfield~\cite{garfield}, we introduce a scale-gate mechanism to split the feature space into different sub-spaces for various 3D physical scales. Then, a 3D Gaussian can belong to different segmentation targets at different granularities without conflict.

\subsubsection{Scale-Gated Affinity Features}
Given a Gaussian affinity feature $\mathbf{f}_{\mathbf{g}}$ and a specific scale $s$, SAGA adopts a scale gate to adapt the magnitude of different feature channels accordingly. The scale gate is defined as a mapping $\mathcal{S}: [0,1] \rightarrow [0,1]^D$, which projects a scale scalar $s \in [0,1]$ to its corresponding soft gate vector $\mathcal{S}(s)$. To maximize segmentation efficiency, the scale gate adopts a extremely streamlined design, which is a single linear layer followed by a sigmoid function. At the scale of $s$ the scale-gated affinity feature is:
\begin{equation}
    \mathbf{f}_{\mathbf{g}}^s = \mathcal{S}(s) \odot \mathbf{f}_{\mathbf{g}},
\end{equation}
where $\odot$ denotes the Hadamard product. Thanks to the simplicity of the scale-gate mechanism, the time overhead caused by scale changing is negligible.

Since all Gaussian affinity features share a common scale gate at scale $s$, during training, we can first render the affinity features to 2D and then 
apply the scale gate to the 2D rendered features, \emph{i.e.},
\begin{equation}
\label{eq:render_feature}
    \mathbf{F}(\mathbf{p})=\sum_{i = 1}^{|\mathcal{G}_{\mathbf{p}}|}\mathbf{f}_{\mathbf{g}_i^{\mathbf{p}}}\alpha_{\mathbf{g}_i^{\mathbf{p}}}\prod_{j=1}^{i-1}(1-\alpha_{\mathbf{g}_j^{\mathbf{p}}}),
\end{equation}
\begin{equation}
    \mathbf{F}^s(\mathbf{p}) = \mathcal{S}(s) \odot \mathbf{F}(\mathbf{p}).
\end{equation}
During inference, the scale gate is directly applied to the 3D Gaussian affinity features for conducting 3D segmentation. 

\paragraph{Local Feature Smoothing}
In practice, we find that there are many noisy Gaussians in the 3D space that exhibit unexpectedly high feature similarities with the segmentation target. This may occur for various reasons, such as insufficient training due to small weights in rasterization or incorrect geometry structure learned by 3D-GS. To tackle this problem, we adopt the spatial locality prior of 3D Gaussians. During training, SAGA uses the smoothed affinity feature of a Gaussian $\mathbf{g}$ to replace its original feature $\mathbf{f}_\mathbf{g}$, \emph{i.e.}, $\mathbf{f}_\mathbf{g} \leftarrow \frac{1}{K}\sum_{\mathbf{g}^\prime \in \texttt{KNN}(\mathbf{g})} \mathbf{f}_{\mathbf{g}^\prime}$. $\texttt{KNN}(\mathbf{g})$ denotes K-nearest neighbors of $\mathbf{g}$. After training, the affinity feature for each 3D Gaussian is saved as its smoothed feature.

\subsection{Scale-Aware Contrastive Learning}
As introduced in Section~\ref{sec:feature}, each 3D Gaussian $\mathbf{g}$ is assigned with an affinity feature $\mathbf{f}_{\mathbf{g}}$. To train these features, we employ a scale-aware contrastive learning strategy to distill the pixel-wise correlation information from 2D masks into 3D Gaussians via the differentiable rasterization. 

\label{sec:training}
\paragraph{Scale-Aware Pixel Identity Vector}
To conduct scale-aware contrastive learning, for an image $\mathbf{I}$, we first convert the automatically extracted 2D masks $\mathcal{M}_{\mathbf{I}}$ to scale-aware supervision signal. For this purpose, we assign a scale-aware pixel identity vector $\mathbf{V}(s, \mathbf{p}) \in \{0,1\}^{N_\mathbf{I}}$ to each pixel $\mathbf{p}$ in $\mathbf{I}$. The identity vectors reflect the 2D masks that a pixel belong to at specific scales. If two pixels $\mathbf{p}_1, \mathbf{p}_2$ share at least a same mask at a given scale (\emph{i.e.}, $\mathbf{V}(s, \mathbf{p}_1) \cdot \mathbf{V}(s, \mathbf{p}_2) > 0$), they should have similar features at scale $s$.

To obtain $\mathbf{V}(s, \mathbf{p})$, we first sort the mask set $\mathcal{M}_{\mathbf{I}}$ in descending order according to their mask scales and get an ordered mask list $\mathcal{O}_{\mathbf{I}} = (\mathbf{M}_{\mathbf{I}}^{(1)}, ..., \mathbf{M}_{\mathbf{I}}^{(N_{\mathbf{I}})})$, where $ s_{\mathbf{M}_{\mathbf{I}}^{(1)}}> ... > s_{\mathbf{M}_{\mathbf{I}}^{(N_{\mathbf{I}})}}$. Then, for a pixel $\mathbf{p}$, when $s_{\mathbf{M}^{(i)}_{\mathbf{I}}} < s$, the i-th entry of $\mathbf{V}(s, \mathbf{p})$ is set to $\mathbf{M}^{(i)}_{\mathbf{I}}(\mathbf{p})$. When $s_{\mathbf{M}^{(i)}_{\mathbf{I}}} \geq s$, the i-th entry of $\mathbf{V}(s, \mathbf{p})$ equals to $1$ only if $\mathbf{M}^{(i)}_{\mathbf{I}}(\mathbf{p}) = 1$ and all smaller masks in $\{ \mathbf{M}^{(j)}_{\mathbf{I}} \mid s \leq s_{\mathbf{M}^{(j)}_{\mathbf{I}}} < s_{\mathbf{M}^{(i)}_{\mathbf{I}}}\}$ equals to $0$ at pixel $\mathbf{p}$. Formally, we have:

\begin{equation}
\begin{aligned}
&\mathbf{V}^i(s, \mathbf{p}) = 
\begin{cases}
\mathbf{M}^{(i)}_{\mathbf{I}}(\mathbf{p}) & \text{if } s_{\mathbf{M}^{(i)}_{\mathbf{I}}} < s \text{ or } \mathcal{C}(\mathbf{p}) \\
0 & \text{otherwise}
\end{cases} \\
&\mathcal{C}(\mathbf{p}) \triangleq \left(\forall \mathbf{M} \in \{\mathbf{M}^{(j)}_{\mathbf{I}} \mid s \leq s_{\mathbf{M}^{(j)}_{\mathbf{I}}} < s_{\mathbf{M}^{(i)}_{\mathbf{I}}}\}, \mathbf{M}(\mathbf{p}) = 0\right)
\end{aligned}
\end{equation}

This assignment of pixel identity vectors is based on the fact that if a pixel belongs to a specific mask at a given scale, it will continue to belong to that mask at larger scales.

\paragraph{Loss Function}
We adapt the correspondence distillation loss~\cite{stego} for training the scale-gated Gaussian affinity features. Concretely, for two pixels $\mathbf{p}_1, \mathbf{p}_2$ at a given scale $s$, their mask correspondence is given by:
\begin{equation}
    \texttt{Corr}_m(s, \mathbf{p}_1,  \mathbf{p}_2) = \mathds{1}(\mathbf{V}(s, \mathbf{p}_1) \cdot \mathbf{V}(s, \mathbf{p}_2)),
\end{equation}
where $\mathds{1}(\cdot)$ is the indicator function, which is equal to 1 when the input greater than or equal to 0. 
The feature correspondence between two pixels is defined as the cosine similarity between their scale-gated features:
\begin{equation}
    \texttt{Corr}_f(s, \mathbf{p}_1, \mathbf{p}_2) = \langle\mathbf{F}^s(\mathbf{p}_1), \mathbf{F}^s(\mathbf{p}_2)\rangle.
\end{equation}
The correspondence distillation loss $\mathcal{L}_{\texttt{corr}}(\mathbf{p}_1, \mathbf{p}_2)$ between two pixels is given by\footnote{The feature correspondence is clipped at 0 to stabilize training. Please refer to~\citet{stego} for more details.}:

\begin{equation}
\begin{aligned}
\mathcal{L}_{\texttt{corr}}(s, \mathbf{p}_1, \mathbf{p}_2) = & \ (1 - 2 \cdot \texttt{Corr}_m(s, \mathbf{p}_1, \mathbf{p}_2)) \\
& \cdot \max(\texttt{Corr}_f(s, \mathbf{p}_1, \mathbf{p}_2), 0)
\end{aligned}
\end{equation}

\paragraph{Feature Norm Regularization} During training, the 2D features are obtained by rendering with 3D affinity features. This indicates a misalignment between 2D and 3D features. As revealed in Equation~\eqref{eq:render_feature}, a 2D feature is a linear combination of multiple 3D features, each with distinct directions. In such situations, SAGA may show good segmentation ability on the rendered feature map but perform poorly in 3D space. This motivates us to introduce a feature norm regularization. Concretely, during rendering the 2D feature map, the 3D features are first normalized as unit vectors, \emph{i.e.},
\begin{equation}
    \mathbf{F}(\mathbf{p})=\sum_{i = 1}^{|\mathcal{G}_{\mathbf{p}}|}
    \frac{\mathbf{f}_{\mathbf{g}_i^{\mathbf{p}}}}{||\mathbf{f}_{\mathbf{g}_i^{\mathbf{p}}}||_2}
    \alpha_{\mathbf{g}_i^{\mathbf{p}}}\prod_{j=1}^{i-1}(1-\alpha_{\mathbf{g}_j^{\mathbf{p}}}).
\end{equation}
Accordingly, $||\mathbf{F}(\mathbf{p})||_2$ ranges in $[0, 1]$. When 3D features along a ray are perfectly aligned, $||\mathbf{F}(\mathbf{p})||_2 = 1$. Thus, we impose a regularization on the rendered feature norm:
\begin{equation}
    \label{eq:rfn}
    \mathcal{L}_{\texttt{norm}}(\mathbf{p}) = 1 - ||\mathbf{F}(\mathbf{p})||_2
\end{equation}

With the feature norm regularization term, for an iteration of training, the loss of SAGA is defined as:
\begin{equation}
    \label{eq:final_loss}
    \begin{aligned}
        \mathcal{L} = \sum_{(\mathbf{p}_1, \mathbf{p}_2) \in \delta(\mathbf{I}) \times \delta(\mathbf{I})} \mathcal{L}_{\texttt{corr}}(\mathbf{p}_1, \mathbf{p}_2) +
        \sum_{\mathbf{p} \in \delta(\mathbf{I})}\mathcal{L}_{\texttt{norm}}(\mathbf{p}), 
    \end{aligned}
\end{equation}
where $\delta(\mathbf{I})$ denotes the set of pixels within the image $\mathbf{I}$.

\subsubsection{Additional Training Strategy}

During the training, an unavoidable issue is the data imbalance, reflected by: 1) Most pixel pairs keep to be positive or negative regardless of scale variations, making the learned feature insensitive to scales; 2) The majority of pixel pairs shows negative correspondence, resulting in feature collapse; 3) Large targets that occupy more pixels in images have more effect on the optimization, leading to bad performance of segmenting small targets. We tackle this problem by resampling the pixel-pairs and re-weighting the loss function for different samples. Please refer to Section A.1 of the appendix.

\subsection{Inference}
\label{sec:infer}

With well-trained Gaussian affinity features, SAGA can conduct various segmentation tasks in the 3D space. For promptable segmentation, SAGA takes \textbf{2D point prompts} at specific view and the scale as input. Then, SAGA segments the 3D target by matching scale-gated 3D Gaussian affinity features with the 2D query features selected from the rendered feature map according to prompt points. For automatic scene decomposing, SAGA employs HDBSCAN to cluster the affinity features directly in the 3D space\footnote{For efficiency, SAGA uniformly selects 1\% of the Gaussians from the 3D-GS model for clustering.}. Additionally, we design a vote-based segmentation mechanism to integrate SAGA with CLIP for conducting open vocabulary segmentation (see Section A.3 of the supplement).

\section{Experiments}
\label{sec:exp}
In this section, we demonstrate the effectiveness of SAGA both quantitatively and qualitatively. For implementation details and experiment settings, please refer to Section A.2.

\subsection{Datasets}
\begin{figure*}[ht]
    \centering
    \includegraphics[width=0.95\linewidth]{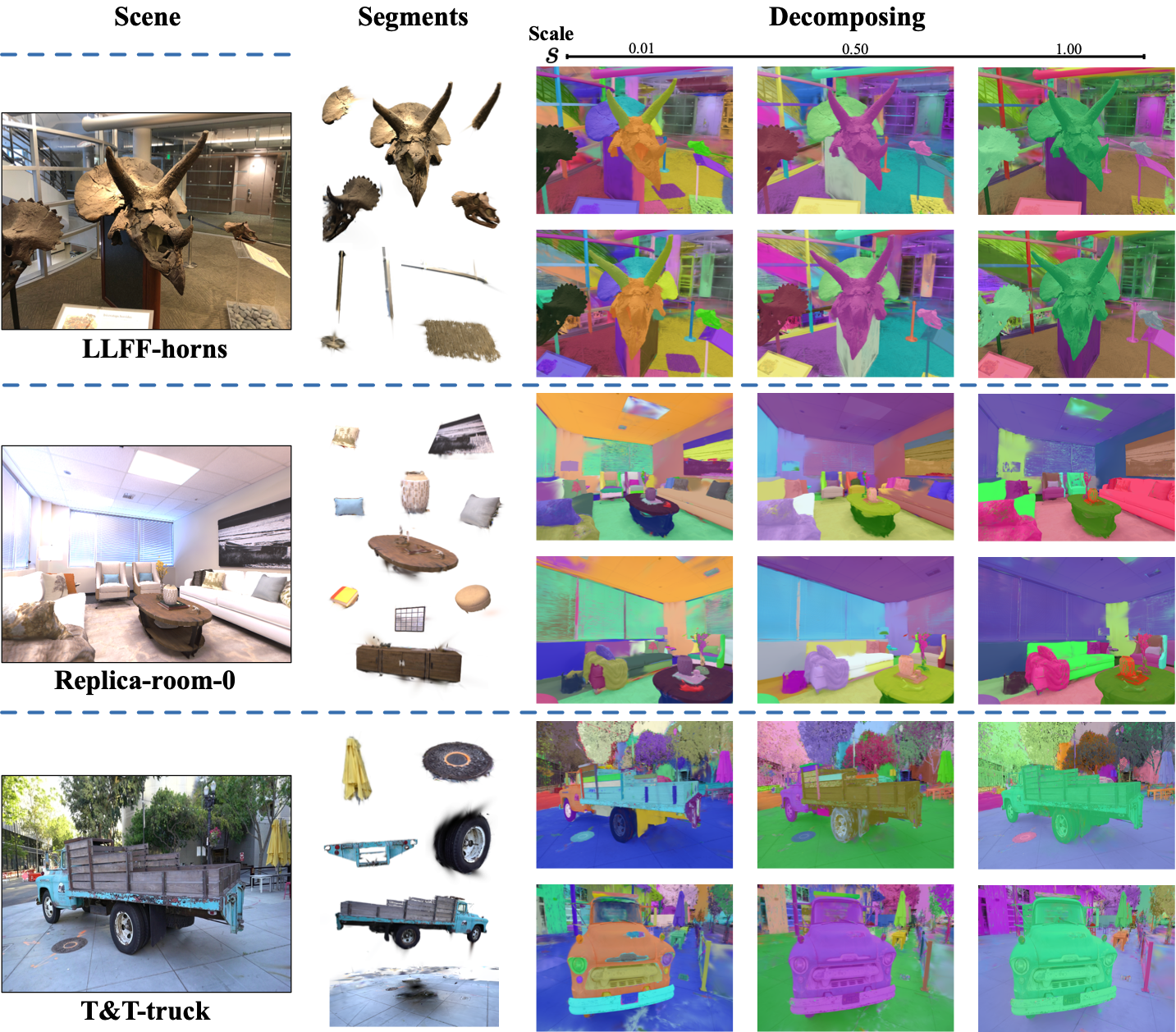}
    \caption{Qualitative results of SAGA across different scenes. We provide both the targets segmented via 2D point prompts and the ``segment everything'' results. }
    \label{fig:vis}
\end{figure*}

For promptable segmentation experiments, we utilize two datasets: NVOS~\cite{nvos} and SPIn-NeRF\cite{spinnerf}. The former is derived from the LLFF dataset~\cite{llff} and the latter is a combination of subsets of data from established NeRF-related datasets~\cite{llff, NeRF, nerf-sup, tanks, plenoxel}. For open-vocabulary segmentation experiments, we adopt the 3D-OVS dataset~\cite{3dovs}. For qualitative analysis (Figure~\ref{fig:vis} and~\ref{fig:hig_frec}), we employ various datasets including LLFF~\cite{llff}, MIP-360~\cite{mipnerf360}, Tanks\&Temple~\cite{tanks}, and Replica~\cite{replica}. These datasets encompass indoor and outdoor scenes, forward-facing and 360-degree scenes, as well as synthetic and real scenes.

\subsection{Quantitative Results}

\paragraph{NVOS Dataset} 
As shown in Table~\ref{tab:nvos}, SAGA outperforms previous segmentation approaches for both 3D-GS and other radiance fields, \textit{i.e.}, +0.4 mIoU over the previous SOTA SA3D-GS and +0.9 mIoU over the OmniSeg3D.

\begin{table}[tbp]
    \centering
    \setlength{\tabcolsep}{1.9mm}{
    \begin{tabular}{lcc}
    \toprule
    Method     & mIoU (\%)& mAcc (\%) \\
    \midrule
    NVOS~\cite{nvos} & 70.1  & 92.0 \\
    ISRF~\cite{isrf} & 83.8 & 96.4 \\
    SA3D~\cite{sa3d} & 90.3 & 98.2  \\
    OmniSeg3D~\cite{omniseg3d} & 91.7 & 98.4  \\
    \midrule
    GauGroup~\cite{gaussiangrouping} & 85.6 & 97.3 \\
    SA3D-GS~\cite{sa3d2} & 92.2 & 98.5  \\
    SAGA (ours) & \textbf{92.6} & \textbf{98.6} \\
    \bottomrule
  \end{tabular}}
 \caption{Results on NVOS dataset.}
 \label{tab:nvos}
\end{table}

\paragraph{SPIn-NeRF Dataset} 
Results on the SPIn-NeRF dataset can be found in Table~\ref{tab:spin}. SA3D performs on par with the previous SOTA OmniSeg3D. The minor performance degradation is attributed to sub-optimal geometry learned by 3D-GS. For example, 3D-GS models the reflection effects with numerous outlier Gaussians that are not aligned with the exact geometry of the object. Excluding these Gaussians results in empty holes in the segmentation mask for certain views, while including them introduces noise in other views. Nevertheless, we believe the segmentation accuracy of SAGA can meet most requirements.

\begin{table}[htbp]
    \centering
    \setlength{\tabcolsep}{1.94mm}{
    \begin{tabular}{lcc}
    \toprule
    Method     & mIoU (\%)& mAcc (\%) \\
    \midrule
    MVSeg~\cite{spinnerf} & 90.9 & 98.9\\
    SA3D~\cite{sa3d} & 92.4 & 98.9  \\
    OmniSeg3D~\cite{omniseg3d} & \textbf{94.3} & \textbf{99.3}  \\
    \midrule
    GauGroup~\cite{gaussiangrouping} & 86.5 & 98.9 \\
    SA-GS~\cite{sags} & 89.9 & 98.7  \\
    SA3D-GS~\cite{sa3d2} & 93.2 & 99.1  \\
    SAGA (ours) & 93.4 & 99.2 \\
    \bottomrule
  \end{tabular}}
 \caption{Results on SPIn-NeRF dataset.}
 \label{tab:spin}
\end{table}

\paragraph{Open-Vocabulary Semantic Segmentation}
As shown in Table~\ref{tab:3d_ovs}, SAGA demonstrates superior results across all scenes in the 3D-OVS dataset. For more details about open-vocabulary segmentation, please refer to Section A.3.

\begin{table}[htbp]
  \centering
  \setlength{\tabcolsep}{1.94mm}{
  \begin{tabular}{lccccc|c}
  \toprule
    Method & bed & bench & room & sofa & lawn & mean\\
    \midrule
    LERF & 73.5 & 53.2 & 46.6 & 27.0 & 73.7 & 54.8\\
    3D-OVS & 89.5 & 89.3 & 92.8 & 74.0 & 88.2 & 86.8\\
    LangSplat & 92.5 & 94.2 & 94.1 & 90.0 & 96.1 & 93.4\\
    N2F2 & 93.8 & 92.6 & 93.5 & 92.1 & 96.3 & 93.9\\
    SAGA & \textbf{97.4} & \textbf{95.4} & \textbf{96.8} & \textbf{93.5} & \textbf{96.6} & \textbf{96.0}\\
    \bottomrule
  \end{tabular}}
  \caption{Results on 3D-OVS dataset (mIoU).}
  \label{tab:3d_ovs}
\end{table}

\subsubsection{Time Consumption Analysis}
\label{sec:time}

In Table~\ref{tab:time}, we reveal the time consumption of SAGA and compare it with existing promptable segmentation methods in radiance fields. ISRF trains a feature field by mimicking the 2D multi-view visual features extracted by DINO~\cite{dino}, thus enjoys faster convergence speed. However, this leads to less accurate segmentation, necessitating extensive post-processing. SA3D and SA-GS employ an iterative mask refinement pipeline, which eliminates the need for training but incurs significant inference time consumption. Compared to methods that distill segmentation capabilities from SAM masks, such as OmniSeg3D and GARField, SAGA demonstrates much faster inference speed and comparable training speed.

\begin{table}[tbp]
  \centering
  \setlength{\tabcolsep}{1.5mm}{
 \begin{tabular}{lcc}
  \toprule
    Method & Training & Inference \\
    \midrule
    SA3D~\cite{sa3d} & - & 45 s \\
    SA-GS~\cite{sags} & - & 15 s \\
    \midrule
    ISRF~\cite{isrf} & 2.5 mins & 3.3 s \\
    OmniSeg3D~\cite{omniseg3d} & 15\textasciitilde 40 mins & 50\textasciitilde 100 ms \\
    GARField~\cite{garfield} & 20\textasciitilde 60 mins & 30\textasciitilde 70 ms \\
    SAGA (ours) & 10\textasciitilde 40 mins & \textbf{2\textasciitilde 5 ms} \\
    \bottomrule
  \end{tabular}}
    \caption{Time consumption comparison.}
    \label{tab:time}
\end{table}

\subsection{Qualitative Results}

\begin{figure}[tbp]
    \centering
    \includegraphics[width=\linewidth]{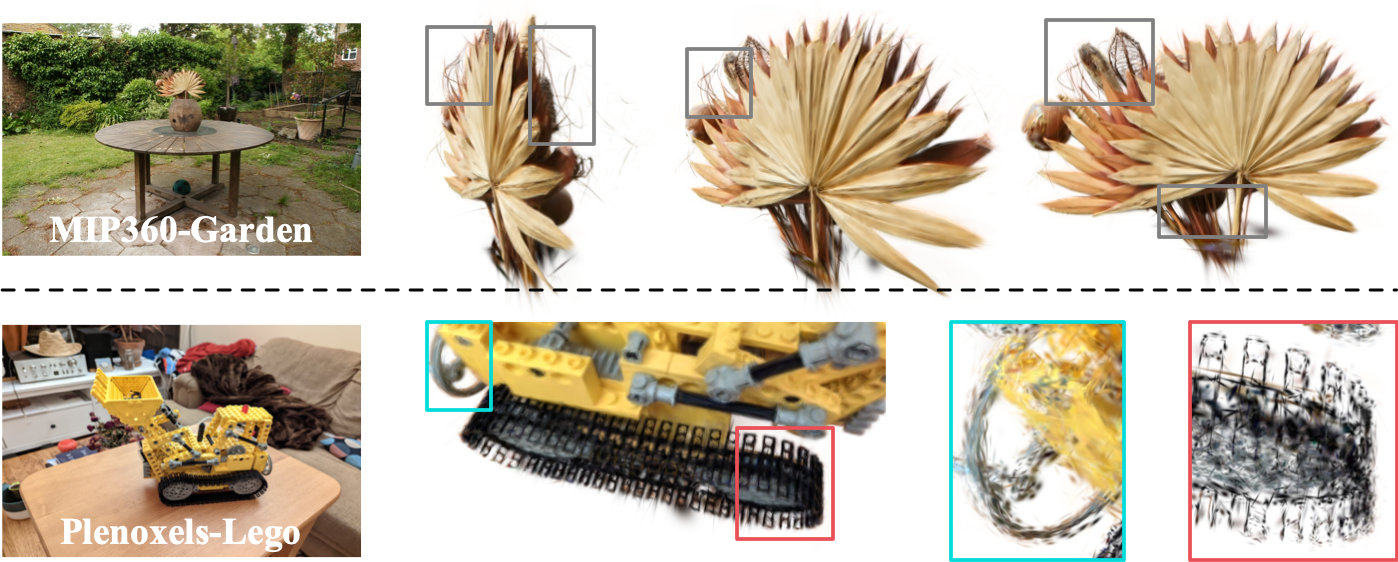}
    \caption{SAGA can maintain the high frequency texture details captured by 3D-GS. We reveal the inherent structure of these details by shrinking the Gaussians by 60\%.}
    \label{fig:hig_frec}
\end{figure}

Figure~\ref{fig:vis} shows that SAGA achieves fine-grained segmentation at various scales across different scenes. The compact Gaussian affinity features enable scene decomposition through simple clustering in the 3D-GS model. The needle-like artifacts on the edges of segmented targets are due to 3D-GS overfitting multi-view RGB without object awareness. Building upon 3D-GS, which can capture high-frequency texture details, SAGA can effectively segment thin, fine-grained structures, as shown in Figure~\ref{fig:hig_frec}. By shrinking the 3D Gaussians, we reveal the underlying structural modeling capabilities of 3D-GS and demonstrate the completeness of the segmentation results.
Since GARField lacks quantitative results on the NVOS and SPIn-NeRF datasets, we provide qualitative comparisons in Section A.6 to highlight the effectiveness of SAGA’s learned features.

\subsection{Ablation Study}
\label{sec:abl}

\paragraph{Local Feature Smoothing (LFS) \& Feature Norm Regularization (FNR)}

Both LFS and FNR impose constraints on the Gaussian affinity features. We present visualization results to illustrate their roles.

\begin{figure}[tbp]
    \centering
    \includegraphics[width=\linewidth]{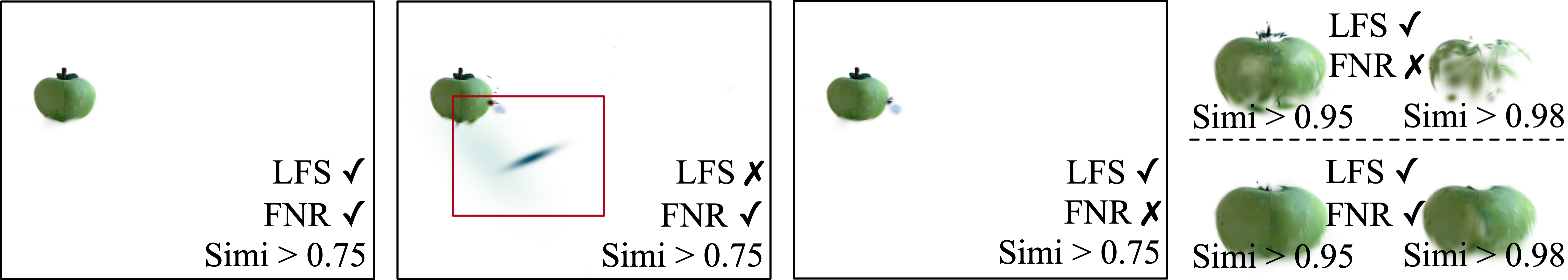}
    \caption{Ablation study on effects of local feature smoothing (Smooth) and feature norm regularization (Feature Norm). Outliers are primarily eliminated through local feature smoothing. Feature norm regularization helps features of inner Gaussians align better with those of the surface.}
    \label{fig:ablation}
\end{figure}

In Figure~\ref{fig:ablation}, when segmenting a 3D object with a cosine similarity threshold of 0.75, without the feature smoothing, the result shows many false positives, which reveals that outliers are primarily eliminated by the feature smoothing operation.
Unlike local feature smoothing, feature norm regularization primarily impacts the Gaussians within objects. When raising the similarity score threshold to 0.95, the apple segmented by SAGA with feature norm regularization remains intact, while the one without it quickly becomes translucent. This phenomenon supports our assumption that 3D features are not perfectly aligned with 2D features, as introduced in Section~\ref{sec:training}. Imposing feature norm regularization helps align the affinity features of 3D Gaussians by pulling the features along a ray in the same direction.

\begin{figure}[tbp]
    \centering
    \includegraphics[width=\linewidth]{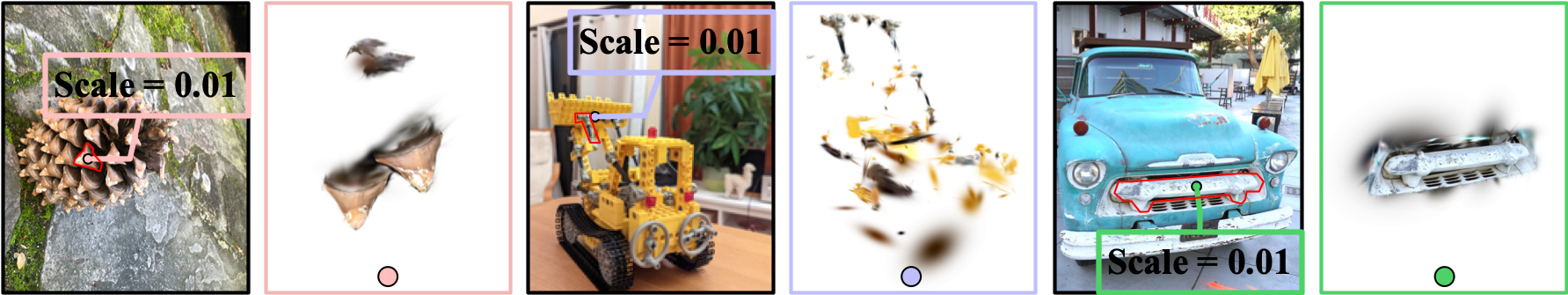}
    \caption{Failure cases of SAGA. The targets of interest are labeled by red border.}
    \label{fig:fail}
\end{figure}

\section{Limitation}
\label{sec:discussion}

SAGA learns the affinity features from multi-view 2D masks extracted by SAM. This makes SAGA hardly segment objects that are not appeared in these masks. As shown in Figure~\ref{fig:fail}, this limitation is particularly evident when the target of interest is small. Enhancing the generalization ability of SAGA to unrecognized targets during the automatic extraction stage is a promising direction.

\section{Conclusion}

In this paper, we propose SAGA, a 3D promptable segmentation method for 3D Gaussian Splatting (3D-GS). SAGA injects the segmentation capability of SAM into Gaussian affinity features for all 3D Gaussians in a 3D-GS model, endowing them with a new property towards segmentation. To preserve the multi-granularity segmentation ability of SAM and the efficiency of 3D-GS, SAGA introduces a lightweight scale-gate mechanism, which adapts the affinity features according to different 3D physical scales with minimal computation overhead. After training, SAGA can achieve real-time fine-grained 3D segmentation. Comprehensive experiments are conducted to demonstrate the effectiveness of SAGA. As one of the first methods addressing promptable segmentation in 3D-GS, the simplicity and effectiveness of SAGA pave the way for future advancements in this field.

\section*{Acknowledgments}
This work was supported by NSFC 62322604, NSFC 62176159, Shanghai Municipal Science and Technology Major Project 2021SHZDZX0102.

\bibliography{aaai25}

\clearpage
\input{appendix}

\end{document}

%% file: appendix.tex
\appendix
\setcounter{figure}{0}
\renewcommand{\thefigure}{A\arabic{figure}}

\section{Appendix}

\begin{figure*}[htbp]
    \centering
    \includegraphics[width=0.9\linewidth]{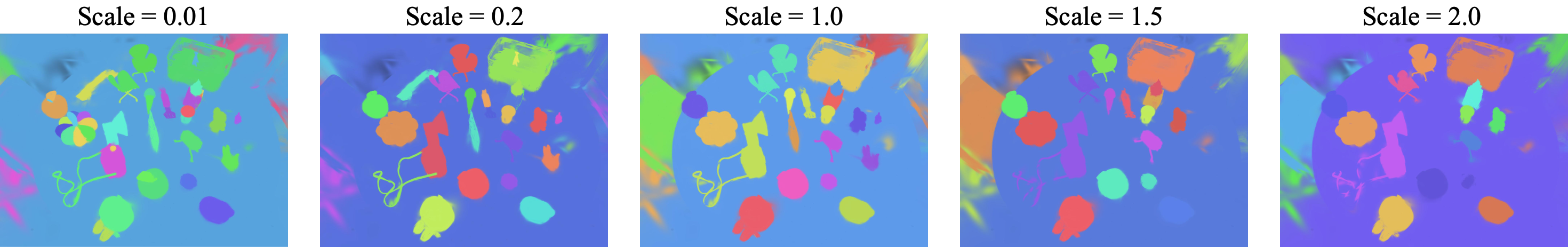}
    \caption{Adapting the scale gate mechanism with GARField achieves competitive results, demonstrating the potential of SAGA across different radiance fields.}
    \label{fig:garfield}
\end{figure*}

In this appendix we provide the concrete training strategy (Section~\ref{sec:train_strategy}) of SAGA and implementation details (Section~\ref{sec:imp_detail}). Then, we provide details about the open-vocabulary segmentation ability of SAGA and analyze its limitation (Section~\ref{sec:lseg}). We also provide an interpretability analysis about the scale gate mechanism (Section~\ref{sec:interpretability}) to reveal the underlying principle of SAGA. Section~\ref{sec:gene} evaluates the robustness and generalizability of SAGA by applying it to more kinds of radiance fields, and Section~\ref{sec:more_vis} presents additional visualization results to demonstrate its effectiveness.

\subsection{Detailed Additional Training Stategy}
\label{sec:train_strategy}

During the contrastive-based learning, an unavoidable problem is the data imbalance. In SAGA, the data imbalance is reflected in the following three aspects: 1) Scale-sensitivity imbalance. The majority of pixel pairs exhibit insensitivity to changes in scale. In other words, during a training iteration, most pixel pairs maintain their positive or negative classification regardless of scale variations. This makes the scale gate collapse to constant output; 2) Positive-negative samples imbalance. The majority of pixel pairs shows negative correspondence, resulting in segmentation features degradation; 3) Target-size imbalance. Large targets that occupy more pixels in images have more effect on the optimization, which leads to bad performance of segmenting small targets.

We use a resampling strategy to tackle the scale-sensitivity imbalance and positive-negative samples imbalance. Then, we adopt a pixel-wise re-weighting strategy to tackle the target-size imbalance.

\paragraph{Resampling} In each iteration of training, we randomly select $N_s$ scales and $N_p$ pixels within an image and form $N_p\times N_p$ pixel pairs. Calculating the mask correspondence for these pixel pairs at every sampled scale results in a scale-conditioned correspondence matrix $\mathbf{R} \in \{0,1\}^{N_sN_pN_p}$. We split the pixel pairs into three sets according to $\mathbf{R}$: 1) Inconsistent set: $\mathcal{Q}^{\texttt{in}} = \{(\mathbf{p}_1, \mathbf{p}_2) \mid \exists s_1,s_2,\ \mathbf{R}(s_1,\mathbf{p}_1, \mathbf{p}_2) \neq \mathbf{R}(s_2,\mathbf{p}_1, \mathbf{p}_2)\}$; 2) Consistent positive set: $\mathcal{Q}^{\texttt{pos}} = \{(\mathbf{p}_1, \mathbf{p}_2) \mid \forall s,\ \mathbf{R}(s,\mathbf{p}_1, \mathbf{p}_2) = 1\}$; 3) Consistent negative set: $\mathcal{Q}^{\texttt{neg}} = \{(\mathbf{p}_1, \mathbf{p}_2) \mid \forall s,\ \mathbf{R}(s,\mathbf{p}_1, \mathbf{p}_2) = 0\}$. During the loss calculation, all pixel pairs in $\mathcal{Q}^{\texttt{in}}$ are involved. Then, we randomly select $\frac{|\mathcal{Q}^{\texttt{in}}|}{2}$ pixel pairs in both $\mathcal{Q}^{\texttt{pos}}$ and $\mathcal{Q}^{\texttt{neg}}$ respectively. To tackle the hard samples in training, we also add pixel pairs in $\mathcal{Q}^{\texttt{neg}}$ which have feature correspondences larger than 0.5 and pixel pairs in $\mathcal{Q}^{\texttt{pos}}$ which have feature correspondence smaller than 0.75 into loss calculation.
This design not only ensures the sensitivity of the loss to scale changing but also keeps the balance of positive pairs and negative pairs.
Let $\phi(\cdot)$ denote the sampling operation, after the resampling, we get three sets of pixel pairs, \emph{i.e.}, $\mathcal{Q}^{\texttt{in}}, \phi(\mathcal{Q}^{\texttt{pos}}), \phi(\mathcal{Q}^{\texttt{neg}})$.

\paragraph{Re-weighting} Considering two masks $\mathbf{M}^1_{\mathbf{I}}, \mathbf{M}^2_{\mathbf{I}}$ in $\mathcal{M}_{\mathbf{I}}$, when uniformly sampling a pair of pixels, the probability that the pair is from $\mathbf{M}^1_{\mathbf{I}}, \mathbf{M}^2_{\mathbf{I}}$ is proportional to the product of the number of positive pixels of the two mask. This indicates that the optimization process is dominated by large masks. To re-weight the loss, we first calculate the mean mask size $m_\mathbf{p} = \frac{1}{|\mathcal{K}_\mathbf{p}|}\sum_{\mathbf{M}\in \mathcal{K}_\mathbf{p}}\sum_{i=1,j=1}^{H,W} \mathbf{M}(i,j)$ for each pixel, where $\mathcal{K}_{\mathbf{p}} = \{\mathbf{M} \mid \mathbf{M}\in \mathcal{M}_{\mathbf{I}}, \mathbf{M}(\mathbf{p}) = 1\}$. For a pixel pair $(\mathbf{p}_1, \mathbf{p}_2)$, the loss weight is defined as $\omega(\mathbf{p}_1, \mathbf{p}_2) = \frac{1}{m_{\mathbf{p}_1}m_{\mathbf{p}_2}}$. All weights in a training iteration are then min-max normalized to the range $[1,10]$ to ensure stable training.

With the resampling, re-weighting strategies, the overall loss function of SAGA is:
\begin{equation}
    \label{eq:final_final_loss}
    \begin{aligned}
        \mathcal{L} = & \frac{\sum_{(\mathbf{p}_1, \mathbf{p}_2) \in \mathcal{Q}^{\texttt{in}} \cup \phi(\mathcal{Q}^{\texttt{pos}})} \omega(\mathbf{p}_1, \mathbf{p}_2) \mathcal{L}_{\texttt{corr}}(\mathbf{p}_1, \mathbf{p}_2)}{|\mathcal{Q}^{\texttt{in}} \cup \phi(\mathcal{Q}^{\texttt{pos}})|} +\\
        &\frac{\sum_{(\mathbf{p}_1, \mathbf{p}_2) \in \mathcal{Q}^{\texttt{in}} \cup \phi(\mathcal{Q}^{\texttt{neg}})} \omega(\mathbf{p}_1, \mathbf{p}_2) \mathcal{L}_{\texttt{corr}}(\mathbf{p}_1, \mathbf{p}_2)}{|\mathcal{Q}^{\texttt{in}} \cup \phi(\mathcal{Q}^{\texttt{neg}})|} +\\
        &\frac{1}{HW}\sum_{\mathbf{p} \in \delta(\mathbf{I})}\mathcal{L}_{\texttt{norm}}(\mathbf{p}), 
    \end{aligned}
\end{equation}
where $\delta(\mathbf{I})$ denotes the set of pixels within the image $\mathbf{I}$.

\subsection{Implementation Details}
\label{sec:imp_detail}

Across different scenes, SAGA maintains consistent hyper-parameters. The feature dimension \(D\) is set to 32. The \(K\) of KNN used in local feature smoothing is set to 16. Training of Gaussian affinity features lasts for 10,000 iterations. In each iteration, we randomly sample eight different scales and 1,000 pixels (\(1000^2\) pixel pairs) from the view for training. For different loss terms, we do not adjust any loss balance coefficients in experiments.

We extract the multi-view 2D masks with the SAM ViT-H model. For open-vocabulary segmentation, we use the OpenCLIP ViT-B/16 model. All training and inference is conducted on a single Nvidia RTX 3090 GPU.

\paragraph{Experiment Setting}

For the NVOS dataset, we randomly select positive and negative points from the scribbles on the reference view (provided by the NVOS dataset) to conduct promptable 3D segmentation. We then render the 3D segmentation result on the target view and evaluate the Intersection over Union (IoU) and pixel-wise accuracy against the provided ground truth. For each scene in the SPIn-NeRF dataset, we randomly select a subset of points within and outside the mask of the reference view as positive and negative prompts.



\subsection{Vote-based Open-vocabulary Segmentation}
\label{sec:lseg}



To enable open-vocabulary 3D segmentation in radiance fields, previous studies~\cite{lerf, 3dovs, langsplat, n2f2, clipgs} focus on aligning 3D language feature fields with the visual features extracted by CLIP~\cite{clip} image encoder. Then 3D segmentation can be achieved by querying the language fields with the textual features. The process of training language feature field can be regarded as multi-view feature fusing, which, in essence, is a kind of vote mechanism. This motivate us to see whether SAGA can handle open-vocabulary segmentation with well-trained Gaussian affinity features at minimum modification.

Readers may wonder why we do not use grounding methods like Grounding-DINO~\cite{grounding_dino} to locate the object and convert the location into point prompts, which can then be fed to SAGA. To answer this question, we emphasize the setting of open-vocabulary segmentation. We follow a stricter approach than previous studies, such as SA3D~\cite{sa3d} and GaussianGrouping~\cite{gaussiangrouping}, which rely on Grounded-SAM to convert language prompts to visual prompts in a specific view. Specifying this view introduces additional prior knowledge. In SAGA, users do not need to query a specific view; all that is required for open-vocabulary segmentation is a text prompt. With this idea in mind, we introduce a vote-based open-vocabulary segmentation strategy.

\paragraph{Constructing Vote Graph by Clustering} 

To cluster the multi-view masks, an intuitive way is to cluster their corresponding segmentation features. However, this is infeasible since the segmentation features in SAGA are scale-conditioned. Segmentation features for different masks are probably in different feature subspace. This drives us to find a kind of \textbf{global features} that is consistent across different scales for multi-view masks to enable global clustering.

We propose to use the segmented Gaussians as the global feature. First, we uniformly sample a set of anchor Gaussians $\mathcal{A}$ from $\mathcal{G}$. Then, for a 2D mask $\mathbf{M} \in \mathcal{M}_\mathbf{I}$ with scale $s_{\mathbf{M}}$, we calculate its scale-conditioned feature as\footnote{For brevity, we continue to use $\delta(\cdot)$ to denote the set of positive pixels in 2D masks.}:

\begin{equation}
    \mathbf{f}_\mathbf{M} = \frac{1}{\delta(\mathbf{M})}\sum_{\mathbf{p} \in \delta(\mathbf{M})}\mathbf{F}^{s_\mathbf{M}}(\mathbf{p}).
\end{equation}

Then we compute the similarities between $\mathbf{f}_\mathbf{M}$ and all anchor Gaussians in $\mathcal{A}$ to get a segmentation result $\mathcal{A}_{\mathbf{M}} = \{\mathbf{g} \mid \langle\mathbf{f}_\mathbf{g}^{s_\mathbf{M}}, \mathbf{f}_\mathbf{M}\rangle > \tau, \mathbf{g} \in \mathcal{A}\}$. The distance between two masks $\mathbf{M}_1, \mathbf{M}_2$ is defined as their intersection over union of this 3D segmentation result, \emph{i.e.}, $D(\mathbf{M}_1, \mathbf{M}_2) = \frac{|\mathcal{A}_{\mathbf{M}_1} \cap \mathcal{A}_{\mathbf{M}_2}|}{|\mathcal{A}_{\mathbf{M}_1} \cup \mathcal{A}_{\mathbf{M}_2}|}$. Then we perform HDBSCAN based on this distance map to cluster the 2D masks.

\paragraph{Vote-based Segmentation}

After clustering, we obtain a vote graph \(\mathcal{V}\), where masks of the same 3D target (instance or part) are grouped together. In other words, each cluster centroid in \(\mathcal{V}\) represents a potential segmentation target in the 3D space. 

For each 2D mask \(\mathbf{M}_\mathbf{I}\) of image \(\mathbf{I}\), we extract its visual feature by feeding the masked image \(\mathbf{I} \odot \mathbf{M}_\mathbf{I}\) to the CLIP visual encoder. During inference, given any text prompt, a relevancy score \(r_\mathbf{M}\) is assigned to each mask \(\mathbf{M}\) by comparing the textual feature with the visual feature of the mask\footnote{We adopt the relevancy score introduced in LERF~\cite{lerf}, which has been proven robust.}. For a cluster centroid \(\mathbf{T}\) in \(\mathcal{V}\), the relevancy scores of its corresponding 2D masks are aggregated to form its final relevancy score, \emph{i.e.},

\begin{equation}
    r_{\mathbf{T}} = \frac{1}{|\mathcal{V}_\mathbf{T}|} \sum_{\mathbf{M} \in \mathcal{V}_\mathbf{T}} r_{\mathbf{M}},
\end{equation}

where \(\mathcal{V}_\mathbf{T}\) is the set containing all 2D masks corresponding to the cluster centroid \(\mathbf{T}\). For semantic segmentation in the 3D-OVS dataset, which provides a category list for each scene, the label of each cluster is assigned as the category with the highest relevancy score. 

\paragraph{Limitation of SAGA in Open-vocabulary Segmentation}

The vote-based open-vocabulary segmentation strategy encounters difficulties in certain scenarios. For instance, when considering a bowl of noodles with an egg in it, SAGA ideally should distinguish between categories such as the egg, the egg with noodles (contents of the bowl), and the bowl with noodles and egg. However, because "egg" is included in masks at larger scales, CLIP often misclassifies larger objects as "egg." This issue is inherently rooted in the multi-granularity ambiguity in semantics. Addressing this problem is a promising research direction.

Another limitation is common among current CLIP-SAM based methods, such as LangSplat~\cite{langsplat} and N2F2~\cite{n2f2}. Both SAGA and these methods first use SAM to segment the images and then use CLIP to attach semantics to these segments by segmenting the corresponding regions of the image and feeding them to CLIP. However, the effectiveness of the CLIP visual encoder also depends on context. For example, SAM sometimes segments the texture on a wall. Without seeing the entire wall, CLIP struggles to recognize it. This lack of context hinders the ability to ground the segments accurately. This is an important yet currently unexplored issue.

\subsection{Interpretability Analysis}
\label{sec:interpretability}

To better understand the scale gate mechanism, we analyze the weights of the learned scale gates through a statistical analysis across 47 scenes, each containing $32$ scale gates, resulting in a total of $47 \times 32$ entries. Within these entries, 36.1\% (543) of the gates are positive, while 63.9\% (961) are negative, meaning that a typical scene has about 12 positive gates and 20 negative gates. In other words, when larger-scale features are used as input, more gates tend to close, and fewer gates remain open, which aligns with the intuitive understanding that finer-grained segmentation requires more features to effectively capture detailed information.

\subsection{SAGA with Hybrid Radiance Fields}
\label{sec:gene}

Although SAGA is designed for segmentation in 3D-GS, its scale gate mechanism is not confined to a particular type of radiance field. We demonstrate this versatility by adapting SAGA to GARField, replacing the scale-conditioned affinity field with a scale-gated affinity field. As illustrated in Figure~\ref{fig:garfield}, the performance remains competitive. 

\begin{figure*}[htbp]
    \centering
    \includegraphics[width=0.9\textwidth]{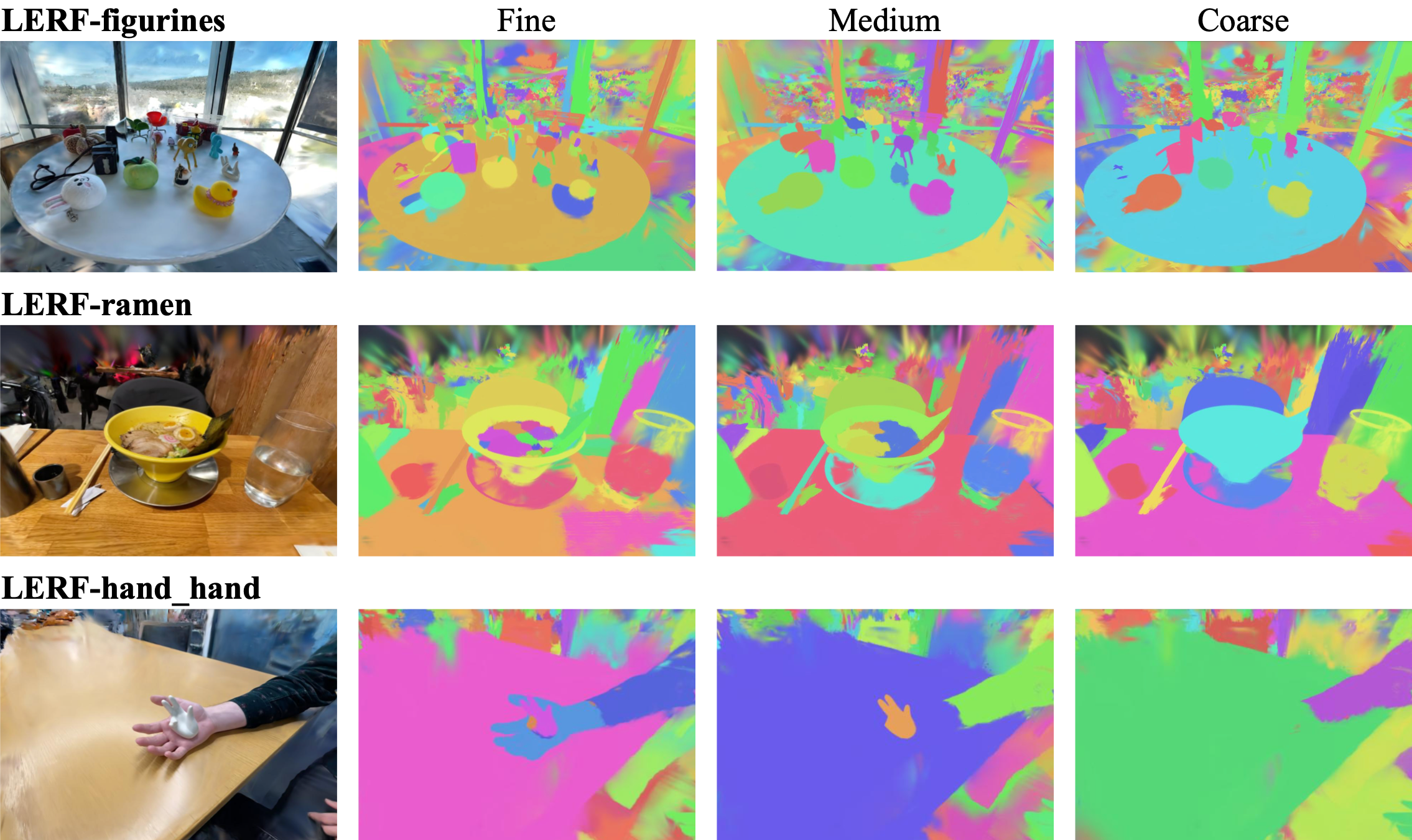}
    \caption{Applying SAGA to Instant-NGP achieves competitive segmentation performance, further demonstrating the generalizability and robustness of SAGA across different radiance field representations.}
    \label{fig:ngp}
\end{figure*}

Furthermore, we apply SAGA to InstantNGP~\cite{InstantNGP}, which can be viewed as an explicit version of GARField without the MLP. As shown in Figure~\ref{fig:ngp}, SAGA continues to produce competitive results. It is worth noting that direct clustering on the hash grid of InstantNGP is infeasible. To address this, we follow GARField’s approach, using 3D-GS to query the hash grids and extract 3D features for clustering.

\subsection{More Qualitative Results}
\label{sec:more_vis}
\begin{figure*}[!htbp]
    \centering
    \includegraphics[width=0.9\linewidth]{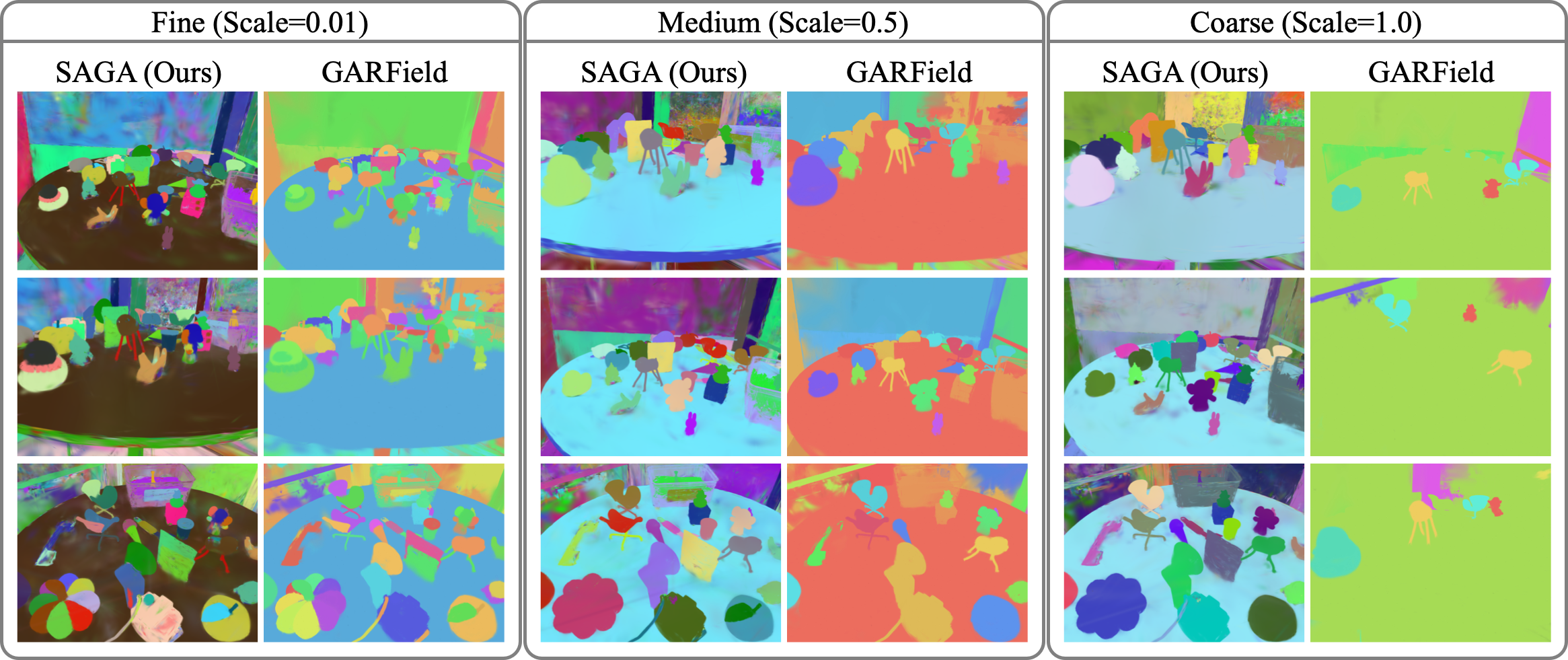}
    \caption{Qualitative comparison with GARField. We conduct feature clustering across the whole scene (LERF-figurines). Compared to GARField, which employs an additional feature field to model affinity features, SAGA demonstrates greater stability by utilizing explicit affinity features. At larger scales, SAGA effectively preserves the perception of small objects without merging them with other targets.}
    \label{fig:comp_garfield}
\end{figure*}

\begin{figure*}[!htbp]
    \centering
    \includegraphics[width=0.9\linewidth]{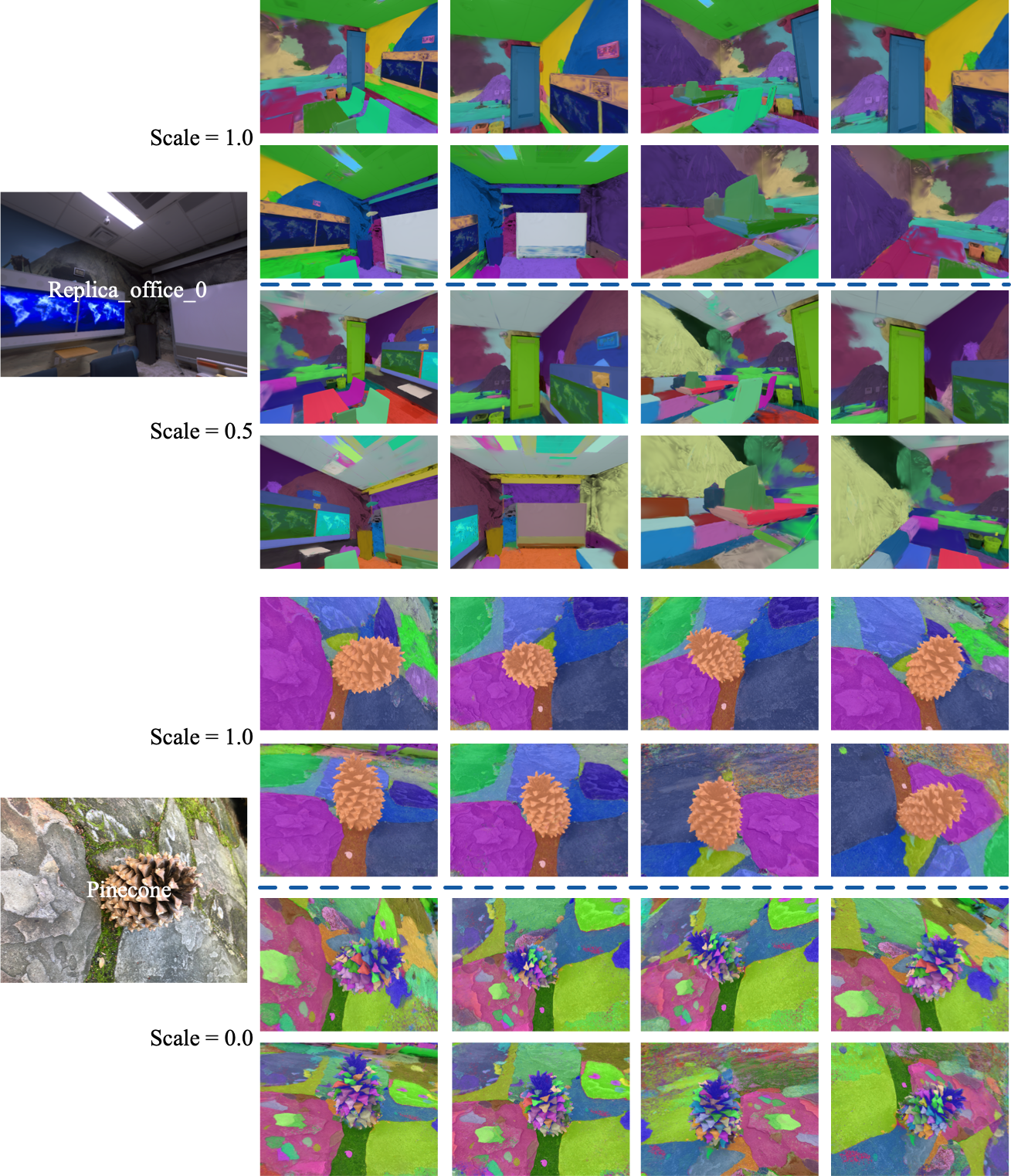}
    \caption{More qualitative results of SAGA.}
    \label{fig:more_vis2}
\end{figure*}

\begin{figure*}[htbp]
    \centering
    \includegraphics[width=0.9\linewidth]{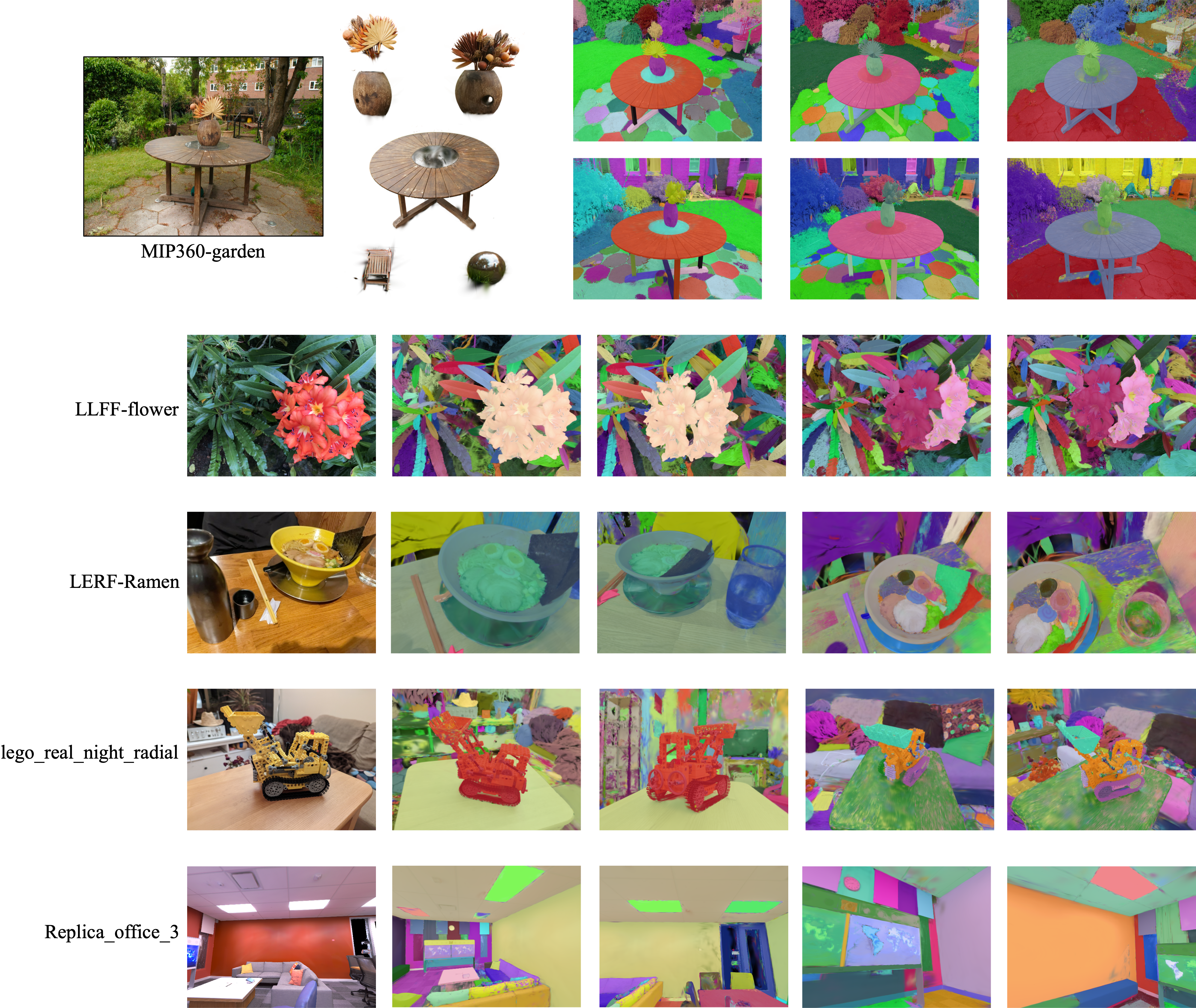}
    \caption{More qualitative results of SAGA.}
    \label{fig:more_vis1}
\end{figure*}

We present additional qualitative results in Figures~\ref{fig:more_vis1} and \ref{fig:more_vis2}. In Figure~\ref{fig:comp_garfield}, it is evident that the affinity features in SAGA exhibit better stability compared to GARField. We perform clustering on the affinity features across the entire scene. At a smaller scale (0.01), both SAGA and GARField achieve fine-grained segmentation. 
However, when conducting coarse segmentation at a larger scale, GARField tends to merge smaller objects into larger ones (\emph{e.g.}, the table). This behavior can be attributed to two factors: first, GARField assumes that the entire scene belongs to the same ``object'' when the scale exceeds that of any existing objects within the scene. Second, GARField employs an implicit feature field to fit the scale-conditioned affinity features. As noted by~\citet{NeRF}, radiance fields often produce over-smoothed predictions, which can result in the loss of small objects at larger scales. In contrast, SAGA assigns an explicit affinity feature to each 3D Gaussian in the 3D-GS model and uses a simple one-layer linear layer (\emph{i.e.}, the scale gate) to model the scale-conditioned effect. This approach allows SAGA to better preserve smaller objects. This is also evidenced by Figure~\ref{fig:garfield}: by replacing the multi-layer perceptron of GARField with our proposed scale-gate mechanism, GARField can preserve the small objects on the table.

\clearpage

%% file: main.bbl
\begin{thebibliography}{63}
\providecommand{\natexlab}[1]{#1}

\bibitem[{Barron et~al.(2022)Barron, Mildenhall, Verbin, Srinivasan, and Hedman}]{mipnerf360}
Barron, J.~T.; Mildenhall, B.; Verbin, D.; Srinivasan, P.~P.; and Hedman, P. 2022.
\newblock Mip-NeRF 360: Unbounded Anti-Aliased Neural Radiance Fields.
\newblock In \emph{CVPR}.

\bibitem[{Bhalgat et~al.(2023)Bhalgat, Laina, Henriques, Vedaldi, and Zisserman}]{contrastivelift}
Bhalgat, Y.; Laina, I.; Henriques, J.~a.~F.; Vedaldi, A.; and Zisserman, A. 2023.
\newblock Contrastive Lift: 3D Object Instance Segmentation by Slow-Fast Contrastive Fusion.
\newblock In \emph{NeurIPS}.

\bibitem[{Bhalgat et~al.(2024)Bhalgat, Laina, Henriques, Zisserman, and Vedaldi}]{n2f2}
Bhalgat, Y.; Laina, I.; Henriques, J.~F.; Zisserman, A.; and Vedaldi, A. 2024.
\newblock N2F2: Hierarchical Scene Understanding with Nested Neural Feature Fields.
\newblock arXiv:2403.10997.

\bibitem[{Bing, Chen, and Yang(2023)}]{dmnerf}
Bing, W.; Chen, L.; and Yang, B. 2023.
\newblock DM-NeRF: 3D Scene Geometry Decomposition and Manipulation from 2D Images.
\newblock In \emph{ICLR}.

\bibitem[{Boykov and Jolly(2001)}]{interactive}
Boykov, Y.~Y.; and Jolly, M.-P. 2001.
\newblock Interactive graph cuts for optimal boundary \& region segmentation of objects in ND images.
\newblock In \emph{ICCV}.

\bibitem[{Caron et~al.(2021)Caron, Touvron, Misra, J{\'{e}}gou, Mairal, Bojanowski, and Joulin}]{dino}
Caron, M.; Touvron, H.; Misra, I.; J{\'{e}}gou, H.; Mairal, J.; Bojanowski, P.; and Joulin, A. 2021.
\newblock Emerging Properties in Self-Supervised Vision Transformers.
\newblock In \emph{ICCV}.

\bibitem[{Cen et~al.(2024)Cen, Fang, Zhou, Yang, Xie, Zhang, Shen, and Tian}]{sa3d2}
Cen, J.; Fang, J.; Zhou, Z.; Yang, C.; Xie, L.; Zhang, X.; Shen, W.; and Tian, Q. 2024.
\newblock Segment Anything in 3D with Radiance Fields.
\newblock arXiv:2304.12308.

\bibitem[{Cen et~al.(2023)Cen, Zhou, Fang, Yang, Shen, Xie, Jiang, Zhang, and Tian}]{sa3d}
Cen, J.; Zhou, Z.; Fang, J.; Yang, C.; Shen, W.; Xie, L.; Jiang, D.; Zhang, X.; and Tian, Q. 2023.
\newblock Segment Anything in 3D with NeRFs.
\newblock In \emph{NeurIPS}.

\bibitem[{Chen et~al.(2022{\natexlab{a}})Chen, Xu, Geiger, Yu, and Su}]{tensorf}
Chen, A.; Xu, Z.; Geiger, A.; Yu, J.; and Su, H. 2022{\natexlab{a}}.
\newblock TensoRF: Tensorial Radiance Fields.
\newblock In \emph{ECCV}.

\bibitem[{Chen et~al.(2023)Chen, Tang, Wan, Wang, and Zeng}]{chen2023interactive}
Chen, X.; Tang, J.; Wan, D.; Wang, J.; and Zeng, G. 2023.
\newblock Interactive Segment Anything NeRF with Feature Imitation.
\newblock arXiv:2305.16233.

\bibitem[{Chen et~al.(2022{\natexlab{b}})Chen, Zhao, Zhang, Duan, Qi, and Zhao}]{chen2022focalclick}
Chen, X.; Zhao, Z.; Zhang, Y.; Duan, M.; Qi, D.; and Zhao, H. 2022{\natexlab{b}}.
\newblock Focalclick: Towards practical interactive image segmentation.
\newblock In \emph{CVPR}.

\bibitem[{Fan et~al.(2023)Fan, Wang, Jiang, Gong, Xu, and Wang}]{nerfsos}
Fan, Z.; Wang, P.; Jiang, Y.; Gong, X.; Xu, D.; and Wang, Z. 2023.
\newblock NeRF-SOS: Any-View Self-supervised Object Segmentation on Complex Scenes.
\newblock In \emph{ICLR}.

\bibitem[{Fang et~al.(2022)Fang, Yi, Wang, Xie, Zhang, Liu, Nie\ss{}ner, and Tian}]{TiNeuVox}
Fang, J.; Yi, T.; Wang, X.; Xie, L.; Zhang, X.; Liu, W.; Nie\ss{}ner, M.; and Tian, Q. 2022.
\newblock Fast Dynamic Radiance Fields with Time-Aware Neural Voxels.
\newblock In \emph{SIGGRAPH Asia 2022 Conference Papers}.

\bibitem[{Fridovich{-}Keil et~al.(2022)Fridovich{-}Keil, Yu, Tancik, Chen, Recht, and Kanazawa}]{plenoxel}
Fridovich{-}Keil, S.; Yu, A.; Tancik, M.; Chen, Q.; Recht, B.; and Kanazawa, A. 2022.
\newblock Plenoxels: Radiance Fields without Neural Networks.
\newblock In \emph{CVPR}.

\bibitem[{Fu et~al.(2022)Fu, Zhang, Chen, Lu, Zhu, Zhou, Geiger, and Liao}]{panopticnerf}
Fu, X.; Zhang, S.; Chen, T.; Lu, Y.; Zhu, L.; Zhou, X.; Geiger, A.; and Liao, Y. 2022.
\newblock Panoptic NeRF: 3D-to-2D Label Transfer for Panoptic Urban Scene Segmentation.
\newblock In \emph{3DV}.

\bibitem[{Goel et~al.(2023)Goel, Sirikonda, Saini, and Narayanan}]{isrf}
Goel, R.; Sirikonda, D.; Saini, S.; and Narayanan, P. 2023.
\newblock Interactive segmentation of radiance fields.
\newblock In \emph{CVPR}.

\bibitem[{Grady(2006)}]{grady2006random}
Grady, L. 2006.
\newblock Random walks for image segmentation.
\newblock \emph{{IEEE} Trans. Pattern Anal. Mach. Intell.}

\bibitem[{Gulshan et~al.(2010)Gulshan, Rother, Criminisi, Blake, and Zisserman}]{gulshan2010geodesic}
Gulshan, V.; Rother, C.; Criminisi, A.; Blake, A.; and Zisserman, A. 2010.
\newblock Geodesic star convexity for interactive image segmentation.
\newblock In \emph{CVPR}.

\bibitem[{Guo et~al.(2024)Guo, Zhu, Peng, Wang, Shen, Hu, and Zhou}]{guo2024samguidedgraphcut3d}
Guo, H.; Zhu, H.; Peng, S.; Wang, Y.; Shen, Y.; Hu, R.; and Zhou, X. 2024.
\newblock SAM-guided Graph Cut for 3D Instance Segmentation.
\newblock arXiv:2312.08372.

\bibitem[{Hamilton et~al.(2022)Hamilton, Zhang, Hariharan, Snavely, and Freeman}]{stego}
Hamilton, M.; Zhang, Z.; Hariharan, B.; Snavely, N.; and Freeman, W.~T. 2022.
\newblock Unsupervised Semantic Segmentation by Distilling Feature Correspondences.
\newblock In \emph{ICLR}.

\bibitem[{Hedman et~al.(2024)Hedman, Srinivasan, Mildenhall, Reiser, Barron, and Debevec}]{bakingnerf}
Hedman, P.; Srinivasan, P.~P.; Mildenhall, B.; Reiser, C.; Barron, J.~T.; and Debevec, P. 2024.
\newblock Baking Neural Radiance Fields for Real-Time View Synthesis.
\newblock \emph{IEEE TPAMI}.

\bibitem[{Hu et~al.(2024)Hu, Wang, Fan, Fan, Peng, Lei, Li, and Zhang}]{sags}
Hu, X.; Wang, Y.; Fan, L.; Fan, J.; Peng, J.; Lei, Z.; Li, Q.; and Zhang, Z. 2024.
\newblock SAGD: Boundary-Enhanced Segment Anything in 3D Gaussian via Gaussian Decomposition.
\newblock arXiv:2401.17857.

\bibitem[{Kerbl et~al.(2023)Kerbl, Kopanas, Leimk{\"u}hler, and Drettakis}]{3dgs}
Kerbl, B.; Kopanas, G.; Leimk{\"u}hler, T.; and Drettakis, G. 2023.
\newblock 3D Gaussian Splatting for Real-Time Radiance Field Rendering.
\newblock \emph{ACM TOG}.

\bibitem[{Kerr et~al.(2023)Kerr, Kim, Goldberg, Kanazawa, and Tancik}]{lerf}
Kerr, J.; Kim, C.~M.; Goldberg, K.; Kanazawa, A.; and Tancik, M. 2023.
\newblock Lerf: Language embedded radiance fields.
\newblock In \emph{ICCV}.

\bibitem[{Kim et~al.(2024)Kim, Wu, Kerr, Tancik, Goldberg, and Kanazawa}]{garfield}
Kim, C.~M.; Wu, M.; Kerr, J.; Tancik, M.; Goldberg, K.; and Kanazawa, A. 2024.
\newblock GARField: Group Anything with Radiance Fields.
\newblock In \emph{CVPR}.

\bibitem[{Kirillov et~al.(2023)Kirillov, Mintun, Ravi, Mao, Rolland, Gustafson, Xiao, Whitehead, Berg, Lo et~al.}]{sam}
Kirillov, A.; Mintun, E.; Ravi, N.; Mao, H.; Rolland, C.; Gustafson, L.; Xiao, T.; Whitehead, S.; Berg, A.~C.; Lo, W.-Y.; et~al. 2023.
\newblock Segment anything.
\newblock In \emph{ICCV}.

\bibitem[{Knapitsch et~al.(2017)Knapitsch, Park, Zhou, and Koltun}]{tanks}
Knapitsch, A.; Park, J.; Zhou, Q.-Y.; and Koltun, V. 2017.
\newblock Tanks and Temples: Benchmarking Large-Scale Scene Reconstruction.
\newblock \emph{{ACM} TOG}.

\bibitem[{Kobayashi, Matsumoto, and Sitzmann(2022)}]{dff}
Kobayashi, S.; Matsumoto, E.; and Sitzmann, V. 2022.
\newblock Decomposing NeRF for Editing via Feature Field Distillation.
\newblock In \emph{NeurIPS}.

\bibitem[{Liao et~al.(2024)Liao, Li, Bao, Ye, Wang, Li, and Liu}]{clipgs}
Liao, G.; Li, J.; Bao, Z.; Ye, X.; Wang, J.; Li, Q.; and Liu, K. 2024.
\newblock CLIP-GS: CLIP-Informed Gaussian Splatting for Real-time and View-consistent 3D Semantic Understanding.
\newblock arXiv:2404.14249.

\bibitem[{Lin et~al.(2022)Lin, Florence, Barron, Lin, Rodriguez, and Isola}]{nerf-sup}
Lin, Y.; Florence, P.; Barron, J.~T.; Lin, T.; Rodriguez, A.; and Isola, P. 2022.
\newblock NeRF-Supervision: Learning Dense Object Descriptors from Neural Radiance Fields.
\newblock In \emph{ICRA}.

\bibitem[{Lindell, Martel, and Wetzstein(2021)}]{autoint}
Lindell, D.~B.; Martel, J. N.~P.; and Wetzstein, G. 2021.
\newblock AutoInt: Automatic Integration for Fast Neural Volume Rendering.
\newblock In \emph{CVPR}.

\bibitem[{Liu et~al.(2023{\natexlab{a}})Liu, Zhan, Zhang, XU, Yu, Saddik, Theobalt, Xing, and Lu}]{3dovs}
Liu, K.; Zhan, F.; Zhang, J.; XU, M.; Yu, Y.; Saddik, A.~E.; Theobalt, C.; Xing, E.; and Lu, S. 2023{\natexlab{a}}.
\newblock Weakly Supervised 3D Open-vocabulary Segmentation.
\newblock In \emph{NeurIPS}.

\bibitem[{Liu et~al.(2023{\natexlab{b}})Liu, Xu, Bertasius, and Niethammer}]{liu2023simpleclick}
Liu, Q.; Xu, Z.; Bertasius, G.; and Niethammer, M. 2023{\natexlab{b}}.
\newblock Simpleclick: Interactive image segmentation with simple vision transformers.
\newblock In \emph{ICCV}.

\bibitem[{Liu et~al.(2024{\natexlab{a}})Liu, Zeng, Ren, Li, Zhang, Yang, Jiang, Li, Yang, Su, Zhu, and Zhang}]{grounding_dino}
Liu, S.; Zeng, Z.; Ren, T.; Li, F.; Zhang, H.; Yang, J.; Jiang, Q.; Li, C.; Yang, J.; Su, H.; Zhu, J.; and Zhang, L. 2024{\natexlab{a}}.
\newblock Grounding DINO: Marrying DINO with Grounded Pre-Training for Open-Set Object Detection.
\newblock arXiv:2303.05499.

\bibitem[{Liu et~al.(2022)Liu, Chen, Yu, Tai, and Tang}]{rfp}
Liu, X.; Chen, J.; Yu, H.; Tai, Y.; and Tang, C. 2022.
\newblock Unsupervised Multi-View Object Segmentation Using Radiance Field Propagation.
\newblock In \emph{NeurIPS}.

\bibitem[{Liu et~al.(2023{\natexlab{c}})Liu, Hu, Huang, Tai, and Tang}]{instance-nerf}
Liu, Y.; Hu, B.; Huang, J.; Tai, Y.-W.; and Tang, C.-K. 2023{\natexlab{c}}.
\newblock Instance neural radiance field.
\newblock In \emph{ICCV}.

\bibitem[{Liu et~al.(2024{\natexlab{b}})Liu, Hu, Tang, and Tai}]{samnerfhq}
Liu, Y.; Hu, B.; Tang, C.-K.; and Tai, Y.-W. 2024{\natexlab{b}}.
\newblock SANeRF-HQ: Segment Anything for NeRF in High Quality.
\newblock In \emph{CVPR}.

\bibitem[{Lyu et~al.(2024)Lyu, Li, Kundu, Tsai, and Yang}]{gaga}
Lyu, W.; Li, X.; Kundu, A.; Tsai, Y.-H.; and Yang, M.-H. 2024.
\newblock Gaga: Group Any Gaussians via 3D-aware Memory Bank.
\newblock arXiv:2404.07977.

\bibitem[{Mildenhall et~al.(2019)Mildenhall, Srinivasan, Cayon, Kalantari, Ramamoorthi, Ng, and Kar}]{llff}
Mildenhall, B.; Srinivasan, P.~P.; Cayon, R.~O.; Kalantari, N.~K.; Ramamoorthi, R.; Ng, R.; and Kar, A. 2019.
\newblock Local light field fusion: practical view synthesis with prescriptive sampling guidelines.
\newblock \emph{{ACM} TOG}.

\bibitem[{Mildenhall et~al.(2020)Mildenhall, Srinivasan, Tancik, Barron, Ramamoorthi, and Ng}]{NeRF}
Mildenhall, B.; Srinivasan, P.~P.; Tancik, M.; Barron, J.~T.; Ramamoorthi, R.; and Ng, R. 2020.
\newblock NeRF: Representing Scenes as Neural Radiance Fields for View Synthesis.
\newblock In \emph{ECCV}.

\bibitem[{Mirzaei et~al.(2023)Mirzaei, Aumentado-Armstrong, Derpanis, Kelly, Brubaker, Gilitschenski, and Levinshtein}]{spinnerf}
Mirzaei, A.; Aumentado-Armstrong, T.; Derpanis, K.~G.; Kelly, J.; Brubaker, M.~A.; Gilitschenski, I.; and Levinshtein, A. 2023.
\newblock {SPIn-NeRF}: Multiview Segmentation and Perceptual Inpainting with Neural Radiance Fields.
\newblock In \emph{CVPR}.

\bibitem[{M{\"{u}}ller et~al.(2022)M{\"{u}}ller, Evans, Schied, and Keller}]{InstantNGP}
M{\"{u}}ller, T.; Evans, A.; Schied, C.; and Keller, A. 2022.
\newblock Instant neural graphics primitives with a multiresolution hash encoding.
\newblock \emph{{ACM} TOG}.

\bibitem[{Niemeyer and Geiger(2021)}]{giraffe}
Niemeyer, M.; and Geiger, A. 2021.
\newblock {GIRAFFE:} Representing Scenes As Compositional Generative Neural Feature Fields.
\newblock In \emph{CVPR}.

\bibitem[{Qin et~al.(2024)Qin, Li, Zhou, Wang, and Pfister}]{langsplat}
Qin, M.; Li, W.; Zhou, J.; Wang, H.; and Pfister, H. 2024.
\newblock LangSplat: 3D Language Gaussian Splatting.
\newblock In \emph{CVPR}.

\bibitem[{Radford et~al.(2021)Radford, Kim, Hallacy, Ramesh, Goh, Agarwal, Sastry, Askell, Mishkin, Clark, Krueger, and Sutskever}]{clip}
Radford, A.; Kim, J.~W.; Hallacy, C.; Ramesh, A.; Goh, G.; Agarwal, S.; Sastry, G.; Askell, A.; Mishkin, P.; Clark, J.; Krueger, G.; and Sutskever, I. 2021.
\newblock Learning Transferable Visual Models From Natural Language Supervision.
\newblock In \emph{ICML}.

\bibitem[{Ren et~al.(2022)Ren, Agarwala, Russell, Schwing, and Wang}]{nvos}
Ren, Z.; Agarwala, A.; Russell, B.~C.; Schwing, A.~G.; and Wang, O. 2022.
\newblock Neural Volumetric Object Selection.
\newblock In \emph{CVPR}.

\bibitem[{Rother, Kolmogorov, and Blake(2004)}]{GrabCut}
Rother, C.; Kolmogorov, V.; and Blake, A. 2004.
\newblock "GrabCut": interactive foreground extraction using iterated graph cuts.
\newblock \emph{{ACM} TOG}.

\bibitem[{Siddiqui et~al.(2023)Siddiqui, Porzi, Bul{\'o}, M{\"u}ller, Nie{\ss}ner, Dai, and Kontschieder}]{siddiqui2023panoptic}
Siddiqui, Y.; Porzi, L.; Bul{\'o}, S.~R.; M{\"u}ller, N.; Nie{\ss}ner, M.; Dai, A.; and Kontschieder, P. 2023.
\newblock Panoptic lifting for 3d scene understanding with neural fields.
\newblock In \emph{CVPR}.

\bibitem[{Sofiiuk, Petrov, and Konushin(2022)}]{sofiiuk2022reviving}
Sofiiuk, K.; Petrov, I.~A.; and Konushin, A. 2022.
\newblock Reviving iterative training with mask guidance for interactive segmentation.
\newblock In \emph{ICIP}.

\bibitem[{Stelzner, Kersting, and Kosiorek(2021)}]{obsurf}
Stelzner, K.; Kersting, K.; and Kosiorek, A.~R. 2021.
\newblock Decomposing 3D Scenes into Objects via Unsupervised Volume Segmentation.
\newblock arXiv:2104.01148.

\bibitem[{Straub et~al.(2019)Straub, Whelan, Ma, Chen, Wijmans, Green, Engel, Mur-Artal, Ren, Verma, Clarkson, Yan, Budge, Yan, Pan, Yon, Zou, Leon, Carter, Briales, Gillingham, Mueggler, Pesqueira, Savva, Batra, Strasdat, Nardi, Goesele, Lovegrove, and Newcombe}]{replica}
Straub, J.; Whelan, T.; Ma, L.; Chen, Y.; Wijmans, E.; Green, S.; Engel, J.~J.; Mur-Artal, R.; Ren, C.; Verma, S.; Clarkson, A.; Yan, M.; Budge, B.; Yan, Y.; Pan, X.; Yon, J.; Zou, Y.; Leon, K.; Carter, N.; Briales, J.; Gillingham, T.; Mueggler, E.; Pesqueira, L.; Savva, M.; Batra, D.; Strasdat, H.~M.; Nardi, R.~D.; Goesele, M.; Lovegrove, S.; and Newcombe, R. 2019.
\newblock The Replica Dataset: A Digital Replica of Indoor Spaces.
\newblock arXiv:1906.05797.

\bibitem[{Sun, Sun, and Chen(2022)}]{dvgo}
Sun, C.; Sun, M.; and Chen, H. 2022.
\newblock Direct Voxel Grid Optimization: Super-fast Convergence for Radiance Fields Reconstruction.
\newblock In \emph{CVPR}.

\bibitem[{Tschernezki et~al.(2022)Tschernezki, Laina, Larlus, and Vedaldi}]{n3f}
Tschernezki, V.; Laina, I.; Larlus, D.; and Vedaldi, A. 2022.
\newblock Neural Feature Fusion Fields: 3D Distillation of Self-Supervised 2D Image Representations.
\newblock In \emph{3DV}.

\bibitem[{Vora et~al.(2022)Vora, Radwan, Greff, Meyer, Genova, Sajjadi, Pot, Tagliasacchi, and Duckworth}]{nesf}
Vora, S.; Radwan, N.; Greff, K.; Meyer, H.; Genova, K.; Sajjadi, M.~S.; Pot, E.; Tagliasacchi, A.; and Duckworth, D. 2022.
\newblock Nesf: Neural semantic fields for generalizable semantic segmentation of 3d scenes.
\newblock \emph{TMLR}.

\bibitem[{Wizadwongsa et~al.(2021)Wizadwongsa, Phongthawee, Yenphraphai, and Suwajanakorn}]{nex}
Wizadwongsa, S.; Phongthawee, P.; Yenphraphai, J.; and Suwajanakorn, S. 2021.
\newblock NeX: Real-Time View Synthesis With Neural Basis Expansion.
\newblock In \emph{CVPR}.

\bibitem[{Xu et~al.(2023)Xu, Yin, Qiu, Liu, Tong, and Han}]{sampro3d}
Xu, M.; Yin, X.; Qiu, L.; Liu, Y.; Tong, X.; and Han, X. 2023.
\newblock SAMPro3D: Locating SAM Prompts in 3D for Zero-Shot Scene Segmentation.
\newblock arXiv:2311.17707.

\bibitem[{Yang et~al.(2023)Yang, Wu, He, Zhao, and Liu}]{sam3d}
Yang, Y.; Wu, X.; He, T.; Zhao, H.; and Liu, X. 2023.
\newblock SAM3D: Segment Anything in 3D Scenes.
\newblock arXiv:2306.03908.

\bibitem[{Ye et~al.(2024)Ye, Danelljan, Yu, and Ke}]{gaussiangrouping}
Ye, M.; Danelljan, M.; Yu, F.; and Ke, L. 2024.
\newblock Gaussian Grouping: Segment and Edit Anything in 3D Scenes.
\newblock arXiv:2312.00732.

\bibitem[{Yin et~al.(2024)Yin, Liu, Xiao, Cohen-Or, Huang, and Chen}]{yin2024sai3d}
Yin, Y.; Liu, Y.; Xiao, Y.; Cohen-Or, D.; Huang, J.; and Chen, B. 2024.
\newblock Sai3d: Segment any instance in 3d scenes.
\newblock In \emph{CVPR}.

\bibitem[{Ying et~al.(2024)Ying, Yin, Zhang, Wang, Yu, Huang, and Fang}]{omniseg3d}
Ying, H.; Yin, Y.; Zhang, J.; Wang, F.; Yu, T.; Huang, R.; and Fang, L. 2024.
\newblock OmniSeg3D: Omniversal 3D Segmentation via Hierarchical Contrastive Learning.
\newblock In \emph{CVPR}.

\bibitem[{Yu, Guibas, and Wu(2022)}]{uorf}
Yu, H.; Guibas, L.~J.; and Wu, J. 2022.
\newblock Unsupervised Discovery of Object Radiance Fields.
\newblock In \emph{ICLR}.

\bibitem[{Zhi et~al.(2021)Zhi, Laidlow, Leutenegger, and Davison}]{semantic-nerf}
Zhi, S.; Laidlow, T.; Leutenegger, S.; and Davison, A.~J. 2021.
\newblock In-Place Scene Labelling and Understanding with Implicit Scene Representation.
\newblock In \emph{ICCV}.

\bibitem[{Zou et~al.(2023)Zou, Yang, Zhang, Li, Li, Wang, Wang, Gao, and Lee}]{seem}
Zou, X.; Yang, J.; Zhang, H.; Li, F.; Li, L.; Wang, J.; Wang, L.; Gao, J.; and Lee, Y.~J. 2023.
\newblock Segment everything everywhere all at once.
\newblock In \emph{NeurIPS}.

\end{thebibliography}
